\documentclass[conference,10pt]{IEEEtran}
\IEEEoverridecommandlockouts
\usepackage{amsmath,amsfonts,amsthm,amssymb,bbm}
\usepackage{cite}

\usepackage{balance}
\usepackage{mathrsfs}
\usepackage{xcolor}
\usepackage{tikz}
\usepackage{pgfplots}
\usepackage{colortbl}
\usetikzlibrary{positioning,quotes,angles,patterns,intersections,pgfplots.fillbetween,calc}
\usepackage{tkz-euclide}
\usepackage{mathtools}
\usepackage{url}
\usetikzlibrary{decorations.pathmorphing}
\usepackage{enumitem}
\usepackage{stfloats}
\usepackage{multirow}
\usepackage{graphicx}
\pgfplotsset{compat=1.17} 
\usepackage[normalem]{ulem}
\DeclareMathOperator*{\argmax}{arg\,max}

\theoremstyle{plain} 
\newtheorem{theorem}{Theorem}

\theoremstyle{definition} 
\theoremstyle{definition} 
\usetikzlibrary{calc,shapes.geometric}

\title{$(\alpha_D,\alpha_G)$-GANs: Addressing GAN Training Instabilities via Dual Objectives}
 \author{%
   \IEEEauthorblockN{Monica Welfert\IEEEauthorrefmark{1},
                     Kyle Otstot\IEEEauthorrefmark{1},
                     Gowtham R. Kurri\IEEEauthorrefmark{2},
                     and Lalitha Sankar\IEEEauthorrefmark{1}}
     \IEEEauthorblockA{\IEEEauthorrefmark{1}%
                     Arizona State University, USA \texttt{\{mwelfert,kotstot,lalithasankar\}@asu.edu}}
   \IEEEauthorblockA{\IEEEauthorrefmark{2}%
                     IIIT Hyderabad, India, \texttt{\{gowtham.kurri\}@iiit.ac.in}}
\thanks{This work is supported in part by NSF grants CIF-1901243, CIF-1815361, CIF-2007688, CIF-2134256, CIF-2031799, and CIF-1934766. Gowtham R. Kurri was with Arizona State University when this work was done.}
}

\def \extended {1} 
\begin{document}
\maketitle
\begin{abstract} 
 %THIS PAPER IS ELIGIBLE FOR THE STUDENT PAPER AWARD. 
 In an effort to address the training instabilities of GANs, we introduce a class of dual-objective GANs with different value functions (objectives) for the generator (G) and discriminator (D). In particular, we model each objective using  $\alpha$-loss, a tunable classification loss, to obtain $(\alpha_D,\alpha_G)$-GANs, parameterized by $(\alpha_D,\alpha_G)\in (0,\infty]^2$. For sufficiently large number of samples and capacities for G and D, we show that the resulting non-zero sum game simplifies to minimizing an $f$-divergence under  appropriate conditions on $(\alpha_D,\alpha_G)$. In the finite sample and capacity setting, 
 we define estimation error to quantify the gap in the generator's performance relative to the optimal setting with infinite samples and obtain upper bounds on this error, showing it to be order optimal under certain conditions. Finally, we highlight the value of tuning $(\alpha_D,\alpha_G)$ in alleviating training instabilities for the synthetic 2D Gaussian mixture ring and the Stacked MNIST datasets.
\end{abstract}

\section{Introduction}
Generative adversarial networks (GANs) have become a crucial data-driven tool for generating synthetic data. GANs are generative models trained to produce samples from an unknown (real) distribution using a finite number of training data samples. They consist of two modules, a generator G and a discriminator D, parameterized by vectors $\theta\in\Theta\subset \mathbb{R}^{n_g}$ and $\omega\in\Omega\subset\mathbb{R}^{n_d}$, respectively, which play an adversarial game with each other. The generator $G_\theta$ maps noise $Z\sim P_Z$ to a data sample in $\mathcal{X}$ via the mapping $z\mapsto G_\theta(z)$ and aims to mimic data from the real distribution $P_{r}$. The discriminator $D_\omega$ takes as input $x\in\mathcal{X}$ and classifies it as real or generated by computing a score $D_\omega(x)\in[0,1]$ which reflects the probability that $x$ comes from $P_r$ (real) as opposed to $P_{G_\theta}$ (synthetic). For a chosen value function $V(\theta,\omega)$, the adversarial game between G and D can be formulated as a zero-sum min-max problem given by 
\thinmuskip=1mu
\begin{align}\label{eqn:GANgeneral}
    \inf_{\theta\in\Theta}\sup_{\omega\in\Omega} \,V(\theta,\omega). 
\end{align}
Goodfellow \emph{et al.}~\cite{Goodfellow14} introduce the vanilla GAN for which 
% \vspace{-0.1in}
% \thinmuskip=0mu
\thickmuskip=2mu
\medmuskip=0mu
% \begin{align}
% V_\text{VG}(\theta,\omega)
% =\mathbb{E}_{X\sim P_r}[\log{D_\omega(X)}]+\mathbb{E}_{X\sim P_{G_\theta}}[\log{(1-D_\omega(X))}],\label{eq:Goodfellowobj}
% \end{align}
% \begin{align}
% V_\text{VG}&(\theta,\omega) \nonumber \\
% &=\mathbb{E}_{X\sim P_r}[\log{D_\omega(X)}]+\mathbb{E}_{X\sim P_{G_\theta}}[\log{(1-D_\omega(X))}].
% \label{eq:Goodfellowobj}
% \end{align}
\begin{align*}
V_\text{VG}(\theta,\omega) \nonumber =\mathbb{E}_{X\sim P_r}[\log{D_\omega(X)}]+\mathbb{E}_{X\sim P_{G_\theta}}[\log{(1-D_\omega(X))}].
\end{align*}
For this $V_\text{VG}$, they show that when the discriminator class $\{D_\omega\}_{\omega\in\Omega}$ is rich enough, \eqref{eqn:GANgeneral} simplifies to minimizing
%$\inf_{\theta\in\Theta} 2D_{\text{JS}}(P_r||P_{G_\theta})-\log{4}$, where $D_{\text{JS}}(P_r||P_{G_\theta})$ is 
the Jensen-Shannon divergence~\cite{Lin91} between $P_r$ and $P_{G_\theta}$. 
% For any $G_\theta$, this is achieved by a discriminator maximizing \eqref{eq:Goodfellowobj}:
%     \begin{align}
%     D_{\omega^*}(x)=\frac{p_r(x)}{p_r(x)+p_{G_\theta}(x)},
%     \end{align}
% where $p_r$ and $p_{G_\theta}$ are the corresponding densities of the distributions $P_r$ and $P_{G_\theta}$, respectively, with respect to a base measure $dx$ (e.g., Lebesgue measure).

% paragraph about other GANs - f-GANs, WGAN, and dual-objective GANs (where generator and discriminator optimize different objective functions) like LSGAN, RenyiGAN
Various other GANs have been studied in the literature using different value functions, including
% Nowozin \emph{et al.} introduced $f$-GANs minimizing specific $f$-divergences
$f$-divergence based GANs called $f$-GANs~\cite{NowozinCT16}, IPM based GANs~\cite{ArjovskyCB17,sriperumbudur2012empirical,liang2018well}, etc. %The aforementioned GANs all optimize a common value function for both $D_\omega$ and $G_\theta$.
%; there have also been a number of dual-objective GANs proposed, where $D_\omega$ and $G_\theta$ optimize different objectives (e.g., Least Squares GAN (LSGAN)~\cite{Mao_2017_LSGAN}, R\'{e}nyiGAN~\cite{bhatia2021least},?).
% Introduce loss-function perspective and alpha-GAN
Observing that the discriminator is a classifier, recently, Kurri \emph{et al.}~\cite{KurriSS21,kurri-2022-convergence} show that the value function in \eqref{eqn:GANgeneral} can be written using a class probability estimation (CPE) loss $\ell(y,\hat{y})$ whose inputs are the true label $y\in\{0,1\}$ and predictor $\hat{y}\in[0,1]$ (soft prediction of $y$) as
% \thinmuskip=0mu
% \thickmuskip=0mu
% \medmuskip=0mu
\begin{align*}
   V(\theta,\omega) =\mathbb{E}_{X\sim P_r}[-\ell(1,D_\omega(X))]+\mathbb{E}_{X\sim P_{G_\theta}}[-\ell(0,D_\omega(X))].
\end{align*}
% \begin{multline}
%    V(\theta,\omega) \\ =\mathbb{E}_{X\sim P_r}[-\ell(1,D_\omega(X))]+\mathbb{E}_{X\sim P_{G_\theta}}[-\ell(0,D_\omega(X))]\label{eqn:lossfnps1}.
% \end{multline}
% \begin{align}
%   &V(\theta,\omega) \nonumber \\
%   &=\mathbb{E}_{X\sim P_r}[-\ell(1,D_\omega(X))]+\mathbb{E}_{X\sim P_{G_\theta}}[-\ell(0,D_\omega(X))]\label{eqn:lossfnps1}
% \end{align}
Using this approach, they introduce $\alpha$-GAN
using the tunable CPE loss $\alpha$-loss~\cite{sypherd2019tunable,sypherd2022journal}, defined for $\alpha \in(0,\infty]$ as
\begin{align} \label{eq:cpealphaloss}
\ell_\alpha(y,\hat{y})\coloneqq\frac{\alpha}{\alpha-1}\left(1-y\hat{y}^{\frac{\alpha-1}{\alpha}}-(1-y)(1-\hat{y})^{\frac{\alpha-1}{\alpha}}\right).
\end{align}
They show that the $\alpha$-GAN formulation recovers various $f$-divergence based GANs including the Hellinger GAN~\cite{NowozinCT16} ($\alpha=1/2$), the vanilla GAN~\cite{Goodfellow14} ($\alpha=1$), and the Total Variation (TV) GAN~\cite{NowozinCT16} ($\alpha=\infty$). Further, for a large enough discriminator class, the min-max optimization for $\alpha$-GAN in \eqref{eqn:GANgeneral} simplifies to minimizing the Arimoto divergence~\cite{osterreicher1996class,LieseV06}.
% This results for the optimal discriminator of the form
% \begin{align}\label{eqn:optimaldoisc}
%     D_{\omega^*}(x)=\frac{p_r(x)^\alpha}{p_r(x)^\alpha+p_{G_\theta}(x)^\alpha}.
% \end{align}

% Introduce GAN training instabilities (vanishing gradients and mode collapse)/ robustness to hyperparameter tuning (e.g., learning rate and training time) and non-saturating version of vanilla GAN - discuss how tuning alpha 
While each of the abovementioned GANs have distinct advantages, they continue to %notoriously 
suffer from one or more types of training instabilities, including vanishing/exploding gradients, mode collapse, and sensitivity to hyperparameter tuning.  
In \cite{Goodfellow14}, Goodfellow \emph{et al.} note that the generator's objective in the vanilla GAN can \emph{saturate} early in training (due to the use of the sigmoid activation) when D can easily distinguish between the real and synthetic samples, i.e., when the output of D is near zero for all synthetic samples, leading to vanishing gradients. Further, a confident D induces a steep gradient at samples close to the real data, thereby preventing G from learning such samples due to exploding gradients.  To alleviate these, \cite{Goodfellow14} proposes a \emph{non-saturating} (NS) generator objective:
\begin{align}
    V_\text{VG}^\text{NS}(\theta,\omega)=\mathbb{E}_{X\sim P_{G_\theta}}[-\log{D_\omega(X)}].
\end{align}

This NS version of the vanilla GAN %\footnote{now an industry standard} 
may be viewed as involving different objective functions for the two players (in fact, with two versions of the $\alpha=1$ CPE loss, i.e., log-loss, for D and G). However, it continues to suffer from mode collapse \cite{arjovsky2017towards,wiatrak2019stabilizing}. While other dual-objective GANs have also been proposed %, where $D_\omega$ and $G_\theta$ optimize different objectives 
(e.g., Least Squares GAN (LSGAN)~\cite{Mao_2017_LSGAN}, R\'{e}nyiGAN~\cite{bhatia2021least}, NS $f$-GAN \cite{NowozinCT16}, hybrid $f$-GAN \cite{poole2016improved}), few have had success fully addressing training instabilities.
%Building on recent results on robust classifiers using $\alpha$-loss, and noting that the discriminator is a classifier, we present a larger class of dual-objective GANs that allow managing the dual 

%Training instabilities result from multiple factors: vanishing or exploding gradients for one player or the other. 
Recent results have shown that $\alpha$-loss demonstrates desirable gradient behaviors for different $\alpha$ values \cite{sypherd2022journal}. It also assures learning robust classifiers that can reduce the confidence of D (a classifier) thereby allowing G to learn without gradient issues. %to be robust to outliers and noisy inputs \cite{sypherd2021journal}, which in turn can potentially avoid the generator from being stuck. 
To this end, we  
introduce a different $\alpha$-loss objective for each player 
%that allows us 
to address training instabilities. % by allowing each player to be robust to their inputs.
%and avoid  
%by using a different $\alpha$-loss-based value function for each player. 
%that allows addressing the competing requirements of jointly training the discriminator and generator. to 
%we can modulate these instabilities and even avoid them altogether in some
% We argue that the competing requirements of jointly training the discriminator and generator modules can be better addressed by choosing appropriate value functions for each module.
%Such an approach has the potential to enable the two modules to learn robustly without being stymied by the other. %To tis   Noting that the discriminator and generator may need to optimize different objectives 
 %this has the potential to enable the two modules to learn robustly without being stymied by the other. %To tis   Noting that the discriminator and generator may need to optimize different objectives 
%To this end, building on \cite{KurriSS21} and \cite{kurri-2022-convergence}, 
We propose a tunable dual-objective $(\alpha_D,\alpha_G)$-GAN, where the objective functions of D and G are written in terms of $\alpha$-loss with parameters $\alpha_D\in(0,\infty]$ and $\alpha_G\in(0,\infty]$, respectively. Our key contributions are:
\begin{itemize}[leftmargin=*]
    \item  For this non-zero sum game, we show that a Nash equilibrium exists. %, provided that we have a sufficiently large number of samples and the generator and discriminator have sufficient capacity. 
    For appropriate $(\alpha_D,\alpha_G)$ values, we derive the optimal strategies for D and G and prove that for the optimal $D_{\omega^*}$, G minimizes an $f$-divergence and can therefore learn the real distribution $P_r$.
    \item Since $\alpha$-GAN captures various GANs, including the vanilla GAN, it can potentially suffer from vanishing gradients due to a saturation effect. We address this by introducing a non-saturating version of the $(\alpha_D,\alpha_G)$-GAN and present its Nash equilibrium strategies for D and G.
    \item A natural question that arises is how to quantify the theoretical guarantees for dual-objective GANs, specifically for $(\alpha_D,\alpha_G)$-GANs, in terms of their estimation capabilities in the setting of limited capacity models and finite training samples. To this end, we define estimation error for $(\alpha_D,\alpha_G)$-GANs, present an upper bound on the error, and a matching lower bound under additional assumptions.
    \item %Finally, we demonstrate empirically that tuning $\alpha_D$ and $\alpha_G$ eliminates vanishing gradients, significantly reduces exploding gradients, and alleviates mode collapse on a synthetic 2D-ring dataset.
    Finally, we demonstrate empirically that tuning $\alpha_D$ and $\alpha_G$ significantly reduces vanishing and exploding gradients and alleviates mode collapse on a synthetic 2D-ring dataset.
    For the high-dimensional Stacked MNIST dataset, we show that our tunable approach is more robust in terms of mode coverage to the choice of GAN hyperparameters, including number of training epochs and learning rate, relative to both vanilla GAN and LSGAN. 
\end{itemize}

\section{Main Results}
\subsection{$(\alpha_D,\alpha_G)$-GAN}
We first propose a dual-objective $(\alpha_D,\alpha_G)$-GAN with different objective functions for the generator and discriminator. In particular, the discriminator maximizes $V_{\alpha_D}(\theta,\omega)$ while the generator minimizes $V_{\alpha_G}(\theta,\omega)$, where
% \thinmuskip=0mu
% \thickmuskip=0mu
\medmuskip=0mu
% \begin{align}
%     &V_{\alpha_D}(\theta,\omega)\nonumber\\
%     &=\mathbb{E}_{X\sim P_r}[-\ell_{\alpha_D}(1,D_\omega(X))]+\mathbb{E}_{X\sim P_{G_\theta}}[-\ell_{\alpha_D}(0,D_\omega(X))]
% \end{align}
% \begin{align}
%     &V_{\alpha_G}(\theta,\omega)\nonumber\\
%     &=\mathbb{E}_{X\sim P_r}[-\ell_{\alpha_G}(1,D_\omega(X))]+\mathbb{E}_{X\sim P_{G_\theta}}[-\ell_{\alpha_G}(0,D_\omega(X))]\label{eqn:sat-gen-objective}.
% \end{align}
\begin{align}
    &V_{\alpha}(\theta,\omega)\nonumber\\
    &=\mathbb{E}_{X\sim P_r}[-\ell_{\alpha}(1,D_\omega(X))]+\mathbb{E}_{X\sim P_{G_\theta}}[-\ell_{\alpha}(0,D_\omega(X))]\label{eqn:sat-gen-objective},
\end{align}
for $\alpha=\alpha_D,\alpha_G \in (0,\infty]$.
We recover the $\alpha$-GAN \cite{KurriSS21,kurri-2022-convergence} value function when $\alpha_D=\alpha_G=\alpha$. The resulting $(\alpha_D,\alpha_G)$-GAN is given by
\begin{subequations}
\begin{align} 
&\sup_{\omega\in\Omega}V_{\alpha_D}(\theta,\omega) \label{eqn:disc_obj} \\
& \inf_{\theta\in\Theta} V_{\alpha_G}(\theta,\omega)
\label{eqn:gen_obj}.
\end{align}
\label{eqn:alpha_D,alpha_G-GAN}
\end{subequations}
% \vspace{-0.01in}
The following theorem presents the conditions under which the optimal generator learns the real distribution $P_r$ when the discriminator set $\Omega$ is large enough.

\begin{figure}[t]
\centering
\footnotesize
\setlength{\tabcolsep}{1pt}
\begin{tabular}{@{}cc@{}}
  \includegraphics[page=1,width=0.48\linewidth]{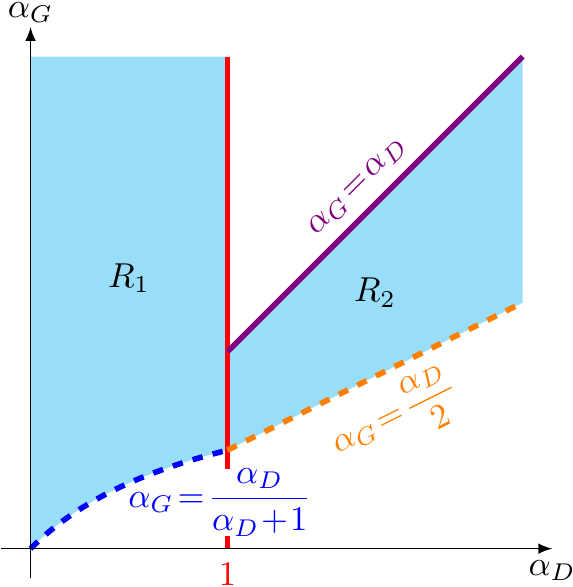}  & \raisebox{1pt}{\includegraphics[page=2,width=0.48\linewidth]{Figures/plots_sat_and_nonsat.pdf}} \\
   (a)  & (b)
\end{tabular}
\caption{(a) Plot of regions $R_1 = \{(\alpha_D,\alpha_G) \in (0,\infty]^2 \bigm\vert \alpha_D \le 1,\alpha_G > \frac{\alpha_D}{\alpha_D+1}\}$ and $R_2 = \{(\alpha_D,\alpha_G) \in (0,\infty]^2 \bigm\vert \alpha_D > 1,\frac{\alpha_D}{2}< \alpha_G \le \alpha_D\}$ for which $f_{\alpha_D,\alpha_G}$ is strictly convex. (b) Plot of region $R_\text{NS}=\{(\alpha_D,\alpha_G)\in (0,\infty]^2 \mid \alpha_D+\alpha_G > \alpha_D\alpha_G\}$ for which $f^\text{NS}_{\alpha_D,\alpha_G}$ is strictly convex.}
\label{fig:convexity_regions}
\end{figure}

\begin{theorem}\label{thm:alpha_D,alpha_G-GAN-saturating}
For a fixed generator $G_\theta$, the discriminator optimizing \eqref{eqn:disc_obj} is given by
\begin{align}
    D_{\omega^*}(x)=\frac{p_r(x)^{\alpha_D}}{p_r(x)^{\alpha_D}+p_{G_\theta}(x)^{\alpha_D}},
    \label{eqn:optimaldisc-gen-alpha-GAN}
\end{align}
where $p_r$ and $p_{G_\theta}$ are the corresponding densities of the distributions $P_r$ and $P_{G_\theta}$, respectively, with respect to a base measure $dx$ (e.g., Lebesgue measure).
For this $D_{\omega^*}$ and the function $f_{\alpha_D,\alpha_G}:\mathbb R_+ \to \mathbb R$ defined as
\begin{align}\label{eqn:f-alpha_d,alpha_g}
f_{\alpha_D,\alpha_G}(u)=\frac{\alpha_G}{\alpha_G-1}\left(\frac{u^{\alpha_D\left(1-\frac{1}{\alpha_G}\right)+1}+1}{(u^{\alpha_D}+1)^{1-\frac{1}{\alpha_G}}}-2^{\frac{1}{\alpha_G}}\right),
\end{align}
\eqref{eqn:gen_obj} simplifies to minimizing a non-negative symmetric $f_{\alpha_D,\alpha_G}$-divergence $D_{f_{\alpha_D,\alpha_G}}(\cdot||\cdot)$ as
\begin{align}\label{eqn:gen-alpha_d,alpha_g-obj}
    \inf_{\theta\in\Theta} D_{f_{\alpha_D,\alpha_G}}(P_r||P_{G_\theta})+\frac{\alpha_G}{\alpha_G-1}\left(2^{\frac{1}{\alpha_G}}-2\right),
\end{align}
which is minimized iff $P_{G_\theta}=P_r$ for $(\alpha_D,\alpha_G)\in (0,\infty]^2$ such that  $\Big (\alpha_D \le 1,\;\alpha_G > \frac{\alpha_D}{\alpha_D+1} \Big ) \; \text{ or } \; \Big ( \alpha_D > 1,\;\frac{\alpha_D}{2}< \alpha_G \le \alpha_D \Big )$. 
% \begin{align*}
%  \Big (\alpha_D \le 1,\;\alpha_G > \frac{\alpha_D}{\alpha_D+1} \Big ) \; \text{ or } \; \Big ( \alpha_D > 1,\;\frac{\alpha_D}{2}< \alpha_G \le \alpha_D \Big ). 
%  % \label{eq:sat-convex-conditions}
% \end{align*}
% $(\alpha_D,\alpha_G)\in R_1 \cup R_2$, where
% \begin{align}
%     R_1 \coloneqq \Big\{(\alpha_D,\alpha_G) \in (0,\infty]^2 \bigm\vert \alpha_D \le 1,\alpha_G > \frac{\alpha_D}{\alpha_D+1}\Big\}
%     \label{eq:convex-reg1-sat}
% \end{align}
% and 
% \begin{align}
%     R_2 \coloneqq \Big\{(\alpha_D,\alpha_G) \in (0,\infty]^2 \bigm\vert \alpha_D > 1,\frac{\alpha_D}{2}< \alpha_G \le \alpha_D\Big\}.
%     \label{eq:convex-reg2-sat}
% \end{align}
%From \eqref{eqn:gen-alpha_d,alpha_g-obj}, $V_{\alpha_G}(\theta,\omega^*)$ is minimized iff $P_{G_\theta}=P_r$.
\end{theorem}
\if \extended 0%
\begin{proof}[Proof sketch]\let\qed\relax
We substitute the optimal discriminator of \eqref{eqn:disc_obj} into the objective function of \eqref{eqn:gen_obj} and translate it into the form
    % in \eqref{eqn:gen-alpha_d,alpha_g-obj} by finding the appropriate conditions on $\alpha_D$ and $\alpha_G$ for $f_{\alpha_D,\alpha_G}$ to be a strictly convex function. Figure \ref{fig:convexity_regions}(a) illustrates the feasible $(\alpha_D,\alpha_G)$-region. A detailed proof can be found in     \if \extended 0%
    \cite[Appendix~A]{welfert2023towards}.
    % 
    % \fi%
    % \if \extended 1%
    % Appendix \ref{appendix:alpha_D,alpha_G-GAN-saturating}.
    % \fi
\end{proof}

\fi%
\if \extended 1%
\begin{proof}
    We substitute the optimal discriminator of \eqref{eqn:disc_obj} into the objective function of \eqref{eqn:gen_obj} and translate it into the form
    \begin{align}
    \int_\mathcal{X} p_{G_\theta}(x)f_{\alpha_D,\alpha_G}\left(\frac{p_r(x)}{p_{G_\theta}(x)}\right) dx + \frac{\alpha_G}{\alpha_G-1}\left(2^{\frac{1}{\alpha_G}}-2\right).
    \label{eq:gen-obj-with-opt-disc}
    \end{align}
    We then find the conditions on $\alpha_D$ and $\alpha_G$ for $f_{\alpha_D,\alpha_G}$ to be strictly convex so that the first term in \eqref{eq:gen-obj-with-opt-disc} is an $f$-divergence. Figure \ref{fig:convexity_regions}(a) illustrates the feasible $(\alpha_D,\alpha_G)$-region. A detailed proof can be found in
    Appendix \ref{appendix:alpha_D,alpha_G-GAN-saturating}.
\end{proof}
\fi

%     \begin{figure}[t]
% \centering
% \includegraphics[page=1,width=2in]{plots_sat_and_nonsat.pdf}
% \caption{Plot of regions $R_1 = \{(\alpha_D,\alpha_G) \in (0,\infty]^2 \bigm\vert \alpha_D \le 1,\alpha_G > \frac{\alpha_D}{\alpha_D+1}\}$ and $R_2 = \{(\alpha_D,\alpha_G) \in (0,\infty]^2 \bigm\vert \alpha_D > 1,\frac{\alpha_D}{2}< \alpha_G \le \alpha_D\}$ for which $f_{\alpha_D,\alpha_G}$ is strictly convex.}
% \label{fig:convexity_region_sat_sym}
% \end{figure}

Noting that $\alpha$-GAN recovers various well-known GANs, including the vanilla GAN, which is prone to saturation, the $(\alpha_D,\alpha_G)$-GAN formulation using the generator objective function in \eqref{eqn:sat-gen-objective} can similarly saturate early in training, causing vanishing gradients. We therefore propose the following NS alternative to the generator's objective in \eqref{eqn:sat-gen-objective}:
\begin{align}
    V^\text{NS}_{\alpha_G}(\theta,\omega) &= \mathbb{E}_{X\sim P_{G_\theta}}[\ell_{\alpha_G}(1,D_\omega(X))],
    \label{eqn:nonsat-gen-objective}
\end{align}
thereby replacing \eqref{eqn:gen_obj} with
\begin{align}
    \inf_{\theta\in\Theta} V^\text{NS}_{\alpha_G}(\theta,\omega).
\label{eqn:gen_obj_ns}
\end{align}

% Comparing \eqref{eqn:sat-gen-objective} and \eqref{eqn:nonsat-gen-objective}, note that the additional expectation term over $P_r$ in \eqref{eqn:sat-gen-objective} results in the symmetric divergence $D_{f_{\alpha_D,\alpha_G}}$ in Theorem \ref{thm:alpha_D,alpha_G-GAN-saturating}, whereas
Comparing \eqref{eqn:gen_obj} and \eqref{eqn:gen_obj_ns}, note that the additional expectation term over $P_r$ in \eqref{eqn:sat-gen-objective} results in \eqref{eqn:gen_obj} simplifying to a symmetric divergence for $D_{\omega^*}$ in \eqref{eqn:optimaldisc-gen-alpha-GAN}, whereas the single term in \eqref{eqn:nonsat-gen-objective} will result in \eqref{eqn:gen_obj_ns} simplifying to an asymmetric divergence.
The optimal discriminator for this NS game remains the same as in \eqref{eqn:optimaldisc-gen-alpha-GAN}. The following theorem provides the solution to \eqref{eqn:gen_obj_ns} under the assumption that the optimal discriminator can be attained.

\begin{theorem}\label{thm:alpha_D,alpha_G-GAN-nonsaturating}
For the same $D_{\omega^*}$ in \eqref{eqn:optimaldisc-gen-alpha-GAN} and the function $f_{\alpha_D,\alpha_G}^\text{NS}:\mathbb R_+ \to \mathbb R$ defined as
\begin{align}\label{eqn:f-alpha_d,alpha_g-ns}
f^\text{NS}_{\alpha_D,\alpha_G}(u)=\frac{\alpha_G}{\alpha_G-1}\left(2^{\frac{1}{\alpha_G}-1}-\frac{u^{\alpha_D\left(1-\frac{1}{\alpha_G}\right)}}{(u^{\alpha_D}+1)^{1-\frac{1}{\alpha_G}}}\right),
\end{align}
\eqref{eqn:gen_obj} simplifies to minimizing a non-negative asymmetric $f^\text{NS}_{\alpha_D,\alpha_G}$-divergence $D_{f^{\text{NS}}_{\alpha_D,\alpha_G}}(\cdot||\cdot)$ as
\begin{align}\label{eqn:gen-alpha_d,alpha_g-obj-ns}
    \inf_{\theta\in\Theta} D_{f^\text{NS}_{\alpha_D,\alpha_G}}(P_r||P_{G_\theta})+\frac{\alpha_G}{\alpha_G-1}\left(1-2^{\frac{1}{\alpha_G}-1}\right),
\end{align}
which is minimized iff $P_{G_\theta}=P_r$ for $(\alpha_D,\alpha_G) \in (0,\infty]^2$ such that $\alpha_D + \alpha_G > \alpha_G\alpha_D.$
% \begin{align}
%     \alpha_D > \alpha_G(\alpha_D-1).
% \end{align}
% $(\alpha_D,\alpha_G) \in R_\text{NS}$, where
% \begin{align}
%     R_\text{NS}=\{(\alpha_D,\alpha_G) \in (0,\infty]^2 \mid \alpha_D > \alpha_G(\alpha_D-1)\}.
%     \label{eq:convex-reg-nonsat}
% \end{align}
\end{theorem}
% \begin{proof}[Proof sketch]\let\qed\relax
%     Similar to the proof of Theorem \ref{thm:alpha_D,alpha_G-GAN-saturating}, we substitute the optimal discriminator of \eqref{eqn:disc_obj} into the objective function of \eqref{eqn:gen_obj_ns} to obtain
%     \begin{align}
%     \int_\mathcal{X} p_{G_\theta}(x)f^\text{NS}_{\alpha_D,\alpha_G}\left(\frac{p_r(x)}{p_{G_\theta}(x)}\right) dx + \frac{\alpha_G}{\alpha_G-1}\left(1-2^{\frac{1}{\alpha_G}-1}\right).
%     \label{eq:gen-obj-with-opt-disc-ns}
%     \end{align}
%     We then find the conditions on $\alpha_D$ and $\alpha_G$ for $f^\text{NS}_{\alpha_D,\alpha_G}$ to be strictly convex so that the first term in \eqref{eq:gen-obj-with-opt-disc-ns} is an $f$-divergence. Proof details can be found in Appendix \ref{appendix:alpha_D,alpha_G-GAN-nonsaturating}.
% \end{proof}
\vspace{-0.05in}
The proof mimics that of Theorem \ref{thm:alpha_D,alpha_G-GAN-saturating} and is detailed in \if \extended 0%
    \cite[Appendix~B]{welfert2023towards}.
    \fi%
    \if \extended 1%
    Appendix \ref{appendix:alpha_D,alpha_G-GAN-nonsaturating}.
    \fi Figure \ref{fig:convexity_regions}(b) illustrates the feasible $(\alpha_D,\alpha_G)$-region; in contrast to the saturating setting of Theorem \ref{thm:alpha_D,alpha_G-GAN-saturating}, the NS setting constrains $\alpha\le 2$ when  $\alpha_D=\alpha_G=\alpha$. Nonetheless, we later show empirically in Section~\ref{subsec:stacked-mnist} that even tuning over this restricted set provides robustness against hyperparameter choices. %Proof details can be found in Appendix .
% \begin{proof}[Proof sketch]\let\qed\relax
%     The proof is similar to that of Theorem \ref{thm:alpha_D,alpha_G-GAN-saturating}. Figure \ref{fig:convexity_regions}(b) illustrates the feasible $(\alpha_D,\alpha_G)$-region in Theorem~\ref{thm:alpha_D,alpha_G-GAN-nonsaturating}. In contrast to the saturating setting of Theorem \ref{thm:alpha_D,alpha_G-GAN-saturating}, the NS setting constrains $\alpha\le 2$ when  $\alpha_D=\alpha_G=\alpha$. Nonetheless, we later show that even tuning over this restricted set provides robustness against hyperparameter choices. Proof details can be found in Appendix \ref{appendix:alpha_D,alpha_G-GAN-nonsaturating}.
% \end{proof}
% \vspace{-.05in}
\subsection{Estimation Error}
\label{subsec:est-err}
Theorems \ref{thm:alpha_D,alpha_G-GAN-saturating} and \ref{thm:alpha_D,alpha_G-GAN-nonsaturating} assume sufficiently large number of training samples and ample discriminator and generator capacity. However, in practice both the number of training samples and model capacity are usually limited. We consider a setting similar to prior works on generalization and estimation error for GANs (e.g., 
%We now consider the same setting as in 
\cite{JiZL21,kurri-2022-convergence}) with finite training samples $S_x=\{X_1,\dots,X_n\}$ and $S_z=\{Z_1,\dots,Z_m\}$ from $P_r$ and $P_Z$, respectively, and with neural networks chosen as the discriminator and generator models. The sets of samples $S_x$ and $S_z$ induce the empirical real and generated distributions $\hat{P}_r$ and $\hat{P}_{G_\theta}$, respectively. A useful quantity to evaluate the performance of GANs in this setting is that of the estimation error, defined in \cite{JiZL21} as the performance gap of the optimized value function when trained using only finite samples relative to the optimal when the statistics are known. Using this definition, \cite{kurri-2022-convergence} derived upper bounds on this error for $\alpha$-GANs. However, such a definition requires a common value function for both discriminator and generator, and therefore, does not directly apply to the dual-objective setting we consider here. 

Our definition relies on the observation that estimation error inherently captures the effectiveness of the generator (for a corresponding optimal discriminator model) in learning with limited samples. We formalize this intuition below.
% To do so, we begin with the definition 
% Kurri \emph{et al.} \cite{kurri-2022-convergence} define estimation error for CPE-loss GANs (inluding $\alpha$-GAN) as
% \begin{align}\label{eqn:estimation-error-def}
%     d^{(\ell)}_{\mathcal{F}_{nn}}(P_r,\hat{P}_{G_{\hat{\theta}^*}})-\inf_{\theta\in\Theta} d^{(\ell)}_{\mathcal{F}_{nn}}(P_r,P_{G_{\theta}}),
% \end{align}
% where $d^{(\ell)}_{\mathcal{F}_{nn}}$ is a loss-inclusive neural net ($nn$) distance defined as
% \begin{align}
%     d^{(\ell)}_{\mathcal{F}_{nn}}(\hat{P}_r,\hat{P}_{G_\theta})=\sup_{\omega\in\Omega}\left(\mathbb{E}_{X\sim \hat{P}_{r}}\phi \big(D_\omega(X) \big)+\mathbb{E}_{X\sim \hat{P}_{G_\theta}}\psi \big(D_\omega(X) \big)\right),
% \end{align}
% for losses $\phi(\cdot)\coloneqq -\ell(1,\cdot)$ and $\psi(\cdot)\coloneqq -\ell(0,\cdot)$, and
% \begin{align}\label{eqn:training-empirical}
%    \hat{\theta}^* = \inf_{\theta\in\Theta}d^{(\ell)}_{\mathcal{F}_{nn}}(\hat{P}_r,\hat{P}_{G_\theta}).
% \end{align}

%We build upon this approach in order to define estimation error for $(\alpha_D,\alpha_G)$-GANs. 
Since $(\alpha_D,\alpha_G)$-GANs use different objective functions for the discriminator and generator, we start by defining the optimal discriminator ${\omega}^*$ for a generator model $G_\theta$ as
% \begin{align}
%     {\omega}^*(P_r,P_{G_\theta}) \coloneqq \argmax_{\omega \in \Omega} &\Big(\mathbb{E}_{X\sim {P}_r}[-\ell_{\alpha_D}(1,D_\omega(X))]+ \nonumber \\
%     &\mathbb{E}_{X\sim {P}_{G_\theta}}[-\ell_{\alpha_D}(0,D_\omega(X))] \Big)
%     \label{eq:est-err-opt-disc}.
% \end{align}
\begin{align}
    {\omega}^*(P_r,P_{G_\theta}) \coloneqq \argmax_{\omega \in \Omega} \; V_{\alpha_D}(\theta,\omega)\big\rvert_{P_r,P_{G_\theta}},
    \label{eq:est-err-opt-disc}
\end{align}
where the notation $|_{\cdot,\cdot}$ allows us to make explicit the distributions used in the value function. %now make explicit the real and generated distributions the value functions are estimated over in order to formalize the estimation error.
In keeping with the literature where the value function being minimized is referred to as the neural net (NN) distance (since D and G are modeled as neural networks) \cite{AroraGLMZ17,JiZL21,kurri-2022-convergence}, we define the generator's NN distance $d_{\omega^*(P_r,P_{G_\theta})}$ as
\begin{align}
    d_{\omega^*(P_r,P_{G_\theta})}(P_r,{P}_{G_{{\theta}}}) \coloneqq V_{\alpha_G}(\theta,\omega^*(P_r,P_{G_\theta}))\big\rvert_{P_r,P_{G_\theta}}.
    \label{eq:est-err-gen-obj}
\end{align}
The resulting minimization for training the $(\alpha_D,\alpha_G)$-GAN using finite samples is
\begin{align}
    \inf_{\theta\in\Theta} d_{\omega^*(\hat{P}_r,\hat{P}_{G_\theta})}(\hat{P}_r,\hat{P}_{G_{{\theta}}}).
    \label{eq:training-empirical-alpha_d,alpha_g-GAN}
\end{align}
Denoting $\hat{\theta}^*$ as the minimizer of \eqref{eq:training-empirical-alpha_d,alpha_g-GAN}, we define the estimation error for $(\alpha_D,\alpha_G)$-GANs as
\begin{align}
    d_{\omega^*(P_r,P_{G_{\hat{\theta}^*}})}(P_r,{P}_{G_{\hat{\theta}^*}})-\inf_{\theta\in\Theta} d_{\omega^*({P}_r,{P}_{G_\theta})}(P_r,P_{G_{\theta}})
    \label{eq:est-error-def-alpha_d,alpha_g-GAN}.
\end{align}
%where $\hat{\theta}^*$ is the minimizer of \eqref{eq:training-empirical-alpha_d,alpha_g-GAN}.

We use the same notation as in \cite{kurri-2022-convergence}, detailed in the following for easy reference. For $x\in\mathcal{X}\coloneqq\{x\in\mathbb{R}^d:||x||_2\leq B_x\}$ and  $z\in\mathcal{Z}\coloneqq\{z\in\mathbb{R}^p:||z||_2\leq B_z\}$, we model the discriminator and generator as $k$- and $l$-layer neural networks, respectively, with
%such that $D_\omega$ and $G_\theta$ can be written as:
\begin{align}
    D_\omega&:x\mapsto \sigma\left(\mathbf{w}_k^\mathsf{T}r_{k-1}(\mathbf{W}_{d-1}r_{k-2}(\dots r_1(\mathbf{W}_1(x)))\right)\,  \label{eqn:disc-model}\\
    G_\theta&:z\mapsto \mathbf{V}_ls_{l-1}(\mathbf{V}_{l-1}s_{l-2}(\dots s_1(\mathbf{V}_1z))),
\end{align}
where (i) $\mathbf{w}_k$ is a parameter vector of the output layer; (ii) for $i\in[1:k-1]$ and $j\in[1:l]$, $\mathbf{W}_i$ and $\mathbf{V}_j$ are parameter matrices; (iii) $r_i(\cdot)$ and $s_j(\cdot)$ are entry-wise activation functions of layers $i$ and $j$, respectively, i.e., for $\mathbf{a}\in\mathbb{R}^t$, $r_i(\mathbf{a})=\left[r_i(a_1),\dots,r_i(a_t)\right]$ and $s_i(\mathbf{a})=\left[s_i(a_1),\dots,s_i(a_t)\right]$; and (iv) $\sigma(\cdot)$ is the sigmoid function given by $\sigma(p)=1/(1+\mathrm{e}^{-p})$. We assume that each $r_i(\cdot)$ and $s_j(\cdot)$ are $R_i$- and $S_j$-Lipschitz, respectively, and also that they are positive homogeneous, i.e., $r_i(\lambda p)=\lambda r_i(p)$ and $s_j(\lambda p)=\lambda s_j(p)$, for any $\lambda\geq 0$ and $p\in\mathbb{R}$. Finally, as is common in such analysis \cite{neyshabur2015norm,salimans2016weight,golowich2018size,JiZL21}, 
we assume that the Frobenius norms of the parameter matrices are bounded, i.e., $||\mathbf{W}_i||_F\leq M_i$, $i\in[1:k-1]$, $||\mathbf{w}_k||_2\leq M_k$, and $||\mathbf{V}_j||_F\leq N_j$, $j\in[1:l]$. We now present an upper bound on \eqref{eq:est-error-def-alpha_d,alpha_g-GAN} in the following theorem.

\begin{theorem}\label{thm:estimationerror-upperbound-alpha_d,alpha_g-GAN}
In the setting described above, with probability at least $1-2\delta$ over the randomness of training samples $S_x=\{X_i\}_{i=1}^n$ and $S_z=\{Z_j\}_{j=1}^m$, we have
\begin{align}
    &d_{\omega^*(P_r,P_{G_{\hat{\theta}^*}})}(P_r,{P}_{G_{\hat{\theta}^*}})-\inf_{\theta\in\Theta} d_{\omega^*({P}_r,{P}_{G_\theta})}(P_r,P_{G_{\theta}})\nonumber\\
    &\leq \frac{4C_{Q_x}(\alpha_G) B_xU_\omega\sqrt{3k}}{\sqrt{n}}+\frac{4C_{Q_z}(\alpha_G) U_\omega U_\theta B_z\sqrt{3(k+l-1)}}{\sqrt{m}}\nonumber\\
    &\hspace{12pt}+U_\omega\sqrt{\log{\frac{1}{\delta}}}\left(\frac{4C_{Q_x}(\alpha_G) B_x}{\sqrt{2n}}+\frac{4C_{Q_z}(\alpha_G) B_zU_\theta}{\sqrt{2m}}\right), \label{eq:estimationbound}
\end{align}
where the parameters $U_\omega\coloneqq M_k\prod_{i=1}^{k-1}(M_iR_i)$ and $U_\theta\coloneqq N_l\prod_{j=1}^{l-1}(N_jS_j)$, $Q_x\coloneqq U_\omega B_x$, $Q_z\coloneqq U_\omega U_\theta B_z$, and
\begin{align} \label{eq:clipalpha}
C_h(\alpha)\coloneqq\begin{cases}\sigma(h)\sigma(-h)^{\frac{\alpha-1}{\alpha}}, \ &\alpha\in(0,1]\\
    \left(\frac{\alpha-1}{2\alpha-1}\right)^{\frac{\alpha-1}{\alpha}}\frac{\alpha}{2\alpha-1}, &\alpha\in(1,\infty).
    \end{cases}
\end{align}
\end{theorem}

\begin{figure*}[h]
    \centering
    \footnotesize
\setlength{\tabcolsep}{1pt}
\begin{tabular}{@{}cc@{}}
  \includegraphics[height=4.3cm]{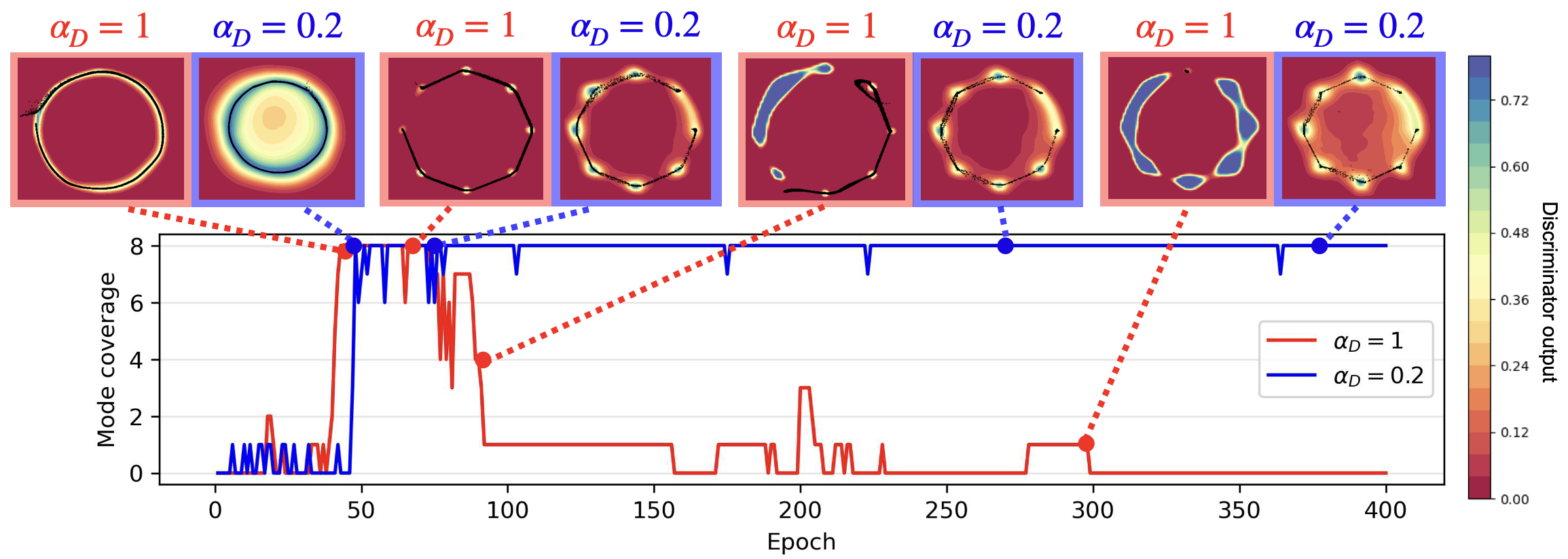}  & \raisebox{1pt}{\includegraphics[height=4.3cm]{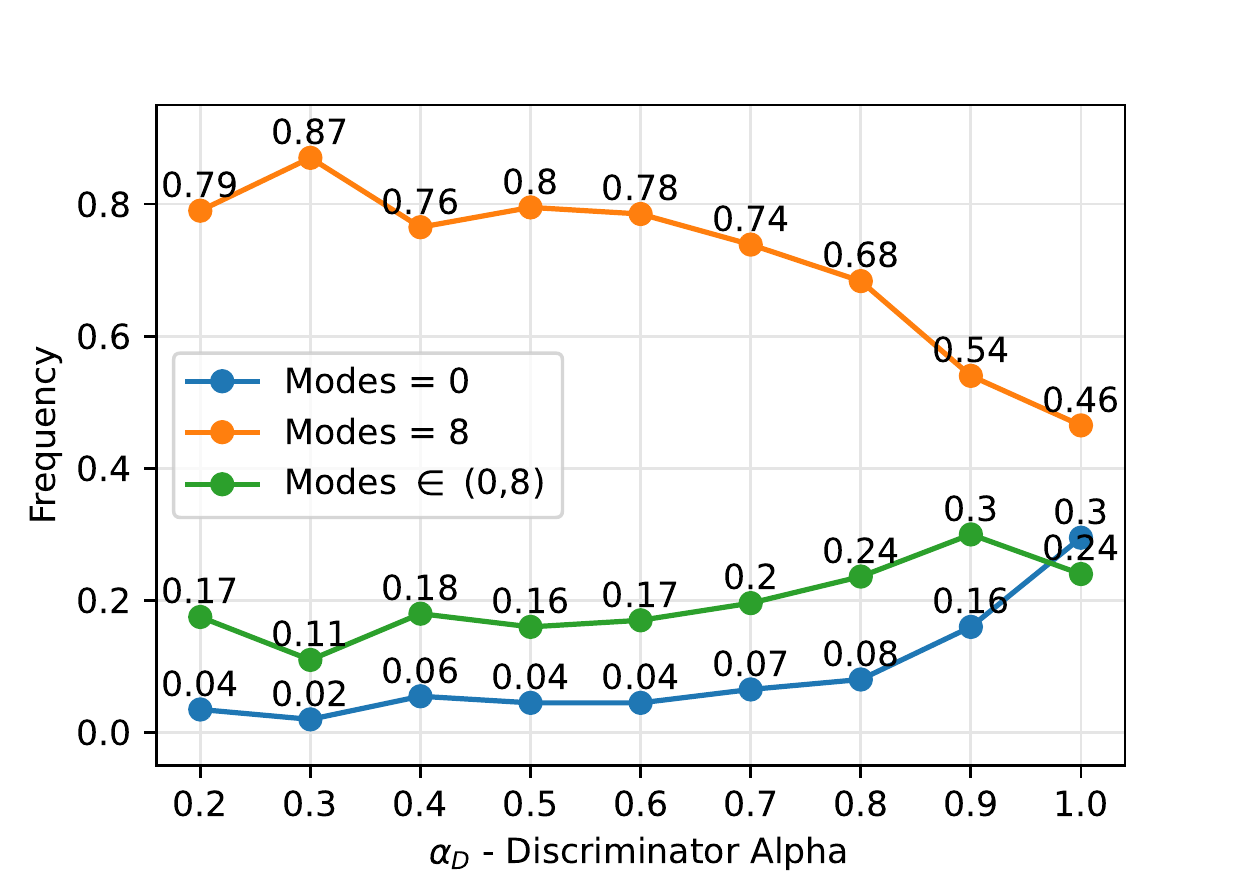}} \\
   (a)  & (b)
\end{tabular}
    \caption{(a) Plot of mode coverage over epochs for $(\alpha_{D}, \alpha_{G})$-GAN training with the \textbf{saturating} objectives in \eqref{eqn:alpha_D,alpha_G-GAN}. Fixing $\alpha_{G}=1$, we compare $\alpha_{D} = 1$ (vanilla GAN) with $\alpha_{D} = 0.2$. Placed above this plot are 2D visuals of the generated samples (in black) at different epochs; these show that both GANs successfully capture the ring-like structure, but the vanilla GAN fails to maintain the ring over time. We illustrate the discriminator output in the same visual as a heat map to show that the $\alpha_{D} = 1$ discriminator exhibits more confident predictions (tending to 0 or 1), which in turn subjects G to vanishing %($\rightarrow 0$) 
    and exploding gradients %($\rightarrow 1$)
    when its objective $\log(1-D)$ saturates as $D\rightarrow 0$ and diverges as $D\rightarrow 1$, respectively. This combination tends to repel the generated data when it approaches the real data, thus freezing any significant weight update in the future. In contrast, the less confident predictions of the $(0.2,1)$-GAN create a smooth landscape for the generated output to descend towards the real data. (b) Plot of success and failure rates over 200 seeds for a range of $\alpha_{D}$ values with $\alpha_{G} = 1$ for the \textbf{saturating} $(\alpha_{D}, \alpha_{G})$-GAN on the 2D-ring, which underscores the stability of $(\alpha_{D} < 1,\alpha_G)$-GANs relative to vanilla GAN.
    }

    \label{fig:sat-figure}
\end{figure*}

% \begin{theorem}\label{thm:estimationerror-upperbound-alpha_d,alpha_g-GAN}
% In the setting described above, with probability at least $1-2\delta$ over the randomness of training samples $S_x=\{X_i\}_{i=1}^n$ and $S_z=\{Z_j\}_{j=1}^m$, we have
% \begin{align}
%     d_{\omega^*(\hat{\theta}^*)}(P_r,{P}_{G_{\hat{\theta}^*}})-\inf_{\theta\in\Theta} d_{\omega^*(\theta)}(P_r,P_{G_{\theta}})\leq K(\alpha_G), \label{eq:estimationbound}
% \end{align}
% where $K(\alpha)$ is defined as the RHS of \cite[eq. (17)]{kurri-2022-convergence} specialized to $\alpha$-loss for $\alpha=\alpha_G$ and the appropriate generator and discriminator model parameters.
% \end{theorem}
    The proof is similar to that of \cite[Theorem 3]{kurri-2022-convergence} (and also \cite[Theorem 1]{JiZL21}). We observe that  
\eqref{eq:estimationbound} does not depend on $\alpha_D$, an artifact of the proof techniques used, and is therefore most likely not the tightest bound possible. See \if \extended 0%
    \cite[Appendix~C]{welfert2023towards}
    \fi%
    \if \extended 1%
    Appendix \ref{appendix:estimationerror-upperbound-alpha_d,alpha_g-GAN}
    \fi for proof details.
% \begin{proof}[Proof sketch]\let\qed\relax
%     The proof is similar to that of \cite[Theorem 3]{kurri-2022-convergence} (and also \cite[Theorem 1]{JiZL21}). We observe that  
% \eqref{eq:estimationbound} does not depend on $\alpha_D$, an artifact of the proof techniques used, and is therefore most likely not the tightest bound possible. See Appendix \ref{appendix:estimationerror-upperbound-alpha_d,alpha_g-GAN} for proof details.
% \end{proof}

% \begin{proof}[Proof sketch]\let\qed\relax
%     We rewrite \eqref{eq:est-error-def-alpha_d,alpha_g-GAN} as a sum of three terms (similarly to \cite{JiZL21,kurri-2022-convergence}) by adding and subtracting relevant terms while keeping the discriminator fixed. We then upper-bound each of the three terms in the same manner as \cite{JiZL21,kurri-2022-convergence}, resulting in the same upper bound as in \cite{kurri-2022-convergence}. See Appendix \ref{appendix:estimationerror-upperbound-alpha_d,alpha_g-GAN} for a detailed proof.
% \end{proof}

When $\alpha_D=\alpha_G=\infty$,  \eqref{eqn:gen-alpha_d,alpha_g-obj} reduces to the total variation distance (up to a constant) \cite[Theorem 2]{KurriSS21}, and \eqref{eq:est-err-gen-obj} simplifies to the loss-inclusive NN distance $d^{\ell}_{\mathcal{F}_{nn}}(\cdot,\cdot)$ defined in \cite[eq. (13)]{kurri-2022-convergence} with $\phi(\cdot)=-\ell_\alpha(1,\cdot)$ and $\psi(\cdot)=-\ell_\alpha(0,\cdot)$ for $\alpha=\infty$. We consider a slightly modified version of this quantity with an added constant to ensure nonnegativity (more details in \if \extended 0%
    \cite[Appendix~D]{welfert2023towards}).
    \fi%
    \if \extended 1%
    Appendix \ref{appendix:est-error-lower-bound-alpha-infinity}).
    \fi % term $1$.  
% \begin{align}
% d^{\ell_\infty}_{\mathcal{F}_{nn}}(P,Q) \coloneqq  \sup_{\omega \in \Omega} \left(\mathbb{E}_{X\sim P}[D_\omega(X) ]+\mathbb{E}_{X\sim Q}\psi \big(D_\omega(X) \big)\right). 
% \end{align}
For brevity, we henceforth denote this as $d^{\ell_\infty}_{\mathcal{F}_{nn}}(\cdot,\cdot)$. 
As in \cite{JiZL21}, suppose the generator's class $\{G_\theta\}_{\theta \in \Theta}$ is rich enough such that the generator $G_\theta$ can learn the real distribution $P_r$ and that the number $m$ of training samples in $S_z$ scales faster than the number $n$ of samples in $S_x$\footnote{Since the noise distribution $P_Z$ is known, one can generate an arbitrarily large number $m$ of noise samples.}. Then $\inf_{\theta \in \Theta} d^{\ell_\infty}_{\mathcal{F}_{nn}}(P_r,P_{G_\theta}) = 0$, so the estimation error simplifies to the single term $d^{\ell_\infty}_{\mathcal{F}_{nn}}(P_r,P_{G_{\hat{\theta}^*}})$. Furthermore, the upper bound in \eqref{eq:estimationbound} reduces to $O(c/\sqrt{n})$ for some constant $c$ (note that, in \eqref{eq:clipalpha}, $C_h(\infty)=1/4$). In addition to the above assumptions, also assume the activation functions $r_i$ for $i \in [1:k-1]$ are either strictly increasing or ReLU. For the above setting, we derive a matching min-max lower bound (up to a constant multiple) on the estimation error.
\begin{theorem}
\label{thm:est-error-lower-bound-alpha-infinity}
For the setting above, let $\hat{P}_n$ be an estimator of $P_r$ learned using the training samples $S_x=\{X_i \}_{i=1}^n$. Then,
\[\inf_{\hat{P}_n} \sup_{P_r \in \mathcal{P}(\mathcal{X})} \, \mathbb P\left\{d^{\ell_\infty}_{\mathcal{F}_{nn}}(\hat{P}_n,P_r) \ge \frac{C(\mathcal{P}(\mathcal{X}))}{\sqrt{n}} \right\} > 0.24,\]
where the constant $C(\mathcal{P}(\mathcal{X}))$ is given by
\begin{align}
    C(\mathcal{P}(\mathcal{X})) = \frac{\log(2)}{20} \Big[ \sigma&(M_k r_{k-1}(\dots r_1(M_1 B_x)) \nonumber \\
    &- \sigma(M_k r_{k-1}(\dots r_1(-M_1 B_x)) \Big].
    \label{eq:est-error-lower-bound-constant}
\end{align}
\end{theorem}

\if \extended 0
\begin{proof}[Proof sketch]\let\qed\relax
    %Oft-used techniques to obtain min-max lower bounds on the quality of an estimator (e.g., LeCam's methods, Fano's methods, etc.) require a semi-metric distance measure, 
    % To obtain min-max  lower bounds, we first prove that $d^{\ell_\infty}_{\mathcal{F}_{nn}}$ is a semi-metric. The remainder of the proof is similar to that of \cite[Theorem 2]{JiZL21}, replacing $d_{\mathcal{F}_{nn}}$ with $d^{\ell_\infty}_{\mathcal{F}_{nn}}$ and noting that the additional sigmoid activation function after the last layer in D satisfies the monotonicity assumption as detailed in \if \extended 0%
    % \cite[Appendix~D]{welfert2023towards}.
    % \fi%
    % \if \extended 1%
    % Appendix \ref{appendix:est-error-lower-bound-alpha-infinity}.
    % \fi A challenge that remains to be addressed is to verify if $d^{\ell_\alpha}_{\mathcal{F}_{nn}}$ is a semi-metric for $\alpha<\infty$.  
     We prove that $d^{\ell_\infty}_{\mathcal{F}_{nn}}$ is a semi-metric. The remainder of the proof is similar to that of \cite[Theorem 2]{JiZL21}. A detailed proof is in \cite[Appendix~D]{welfert2023towards}. .
    %we leave it for later to show that it is a semi-metric. %, as it does not satisfy the triangle inequality. 
\end{proof}
\fi
 \if \extended 1%
\begin{proof}[Proof sketch]\let\qed\relax
    To obtain min-max  lower bounds, we first prove that $d^{\ell_\infty}_{\mathcal{F}_{nn}}$ is a semi-metric. The remainder of the proof is similar to that of \cite[Theorem 2]{JiZL21}, replacing $d_{\mathcal{F}_{nn}}$ with $d^{\ell_\infty}_{\mathcal{F}_{nn}}$ and noting that the additional sigmoid activation function after the last layer in D satisfies the monotonicity assumption as detailed in Appendix \ref{appendix:est-error-lower-bound-alpha-infinity}. A challenge that remains to be addressed is to verify if $d^{\ell_\alpha}_{\mathcal{F}_{nn}}$ is a semi-metric for $\alpha<\infty$.  
\end{proof}
\fi

\section{illustration of Results}

%In this section, we empirically demonstrate the ability of $(\alpha_{D}, \alpha_{G})$-GAN compared to the vanilla GAN as well as the state-of-the-art least squares (LS) GAN \cite{}. Specifically, we evaluate these methods on two datasets: a synthetic dataset generated by a two-dimensional, ring-shaped Gaussian mixture distribution, and the Stacked MNIST image dataset.
In this section, we compare $(\alpha_{D}, \alpha_{G})$-GAN to two state-of-the-art GANs, namely the vanilla GAN  (i.e., the $(1,1)$-GAN) and LSGAN \cite{Mao_2017_LSGAN}, on two datasets: (i) a synthetic dataset generated by a two-dimensional, ring-shaped Gaussian mixture distribution (2D-ring) \cite{srivastava2017veegan} and (ii) the Stacked MNIST image dataset \cite{PACGANLin}. For each dataset and different GAN objectives, we report several metrics that encapsulate the stability of GAN training over hundreds of random seeds. This allows us to clearly %provide insight into our observations which ultimately 
showcase the potential for tuning $(\alpha_{D}, \alpha_{G})$ to obtain stable and robust solutions for image generation.

\subsection{2D Gaussian Mixture Ring}

The 2D-ring is an oft-used synthetic dataset for evaluating GANs. We draw samples from a mixture of 8 equal-prior Gaussian distributions, indexed $i \in \{1,2,\hdots ,  8 \}$ with a mean of $(\cos(2\pi i / 8), \text{ } \sin(2\pi i / 8))$ and variance $10^{-4}$. We generate 50,000 training and 25,000 testing samples; additionally, we generate the same number of 2D latent Gaussian noise vectors.

Both the D and G networks have 4 fully-connected layers with 200 and 400 units, respectively. %The D network has 4 fully-connected layers of 200 units each, while G has 4 fully-connected layers of 400 units each. 
We train for 400 epochs with a batch size of 128, and optimize with Adam \cite{kingma2014adam} and a learning rate of $10^{-4}$ for both models. We consider three distinct settings that differ in the objective functions as: \textbf{(i)} $(\alpha_{D}, \alpha_{G})$-GAN in \eqref{eqn:alpha_D,alpha_G-GAN}; \textbf{(ii)} NS $(\alpha_{D}, \alpha_{G})$-GAN's %\textit{non-saturating} 
in \eqref{eqn:disc_obj}, \eqref{eqn:gen_obj_ns}; \textbf{(iii)} LSGAN with the 0-1 binary coding scheme (see \if \extended 0%
    \cite[Appendix~E]{welfert2023towards}
    \fi%
    \if \extended 1%
    Appendix \ref{appendix:experimental-details-results}
    \fi for details).

For every setting listed above, we train our models on the 2D-ring dataset for 200 random state seeds, where each seed contains different weight initializations for D and G. Ideally, a stable method will reflect similar performance across randomized initializations and also over training epochs; thus, we explore how GAN training performance for each setting varies across seeds and epochs. Our primary performance metric is \textit{mode coverage}, defined as the number of Gaussians (0-8) that contain a generated sample within 3 standard deviations of its mean. A score of 8 conveys successful training, while a score of 0 conveys a significant GAN failure; on the other hand, a score in between 0 and 8 may be indicative of common GAN issues, such as mode collapse or failure to converge. 

\begin{figure}[t]
    \centering
    \includegraphics[width=8cm]{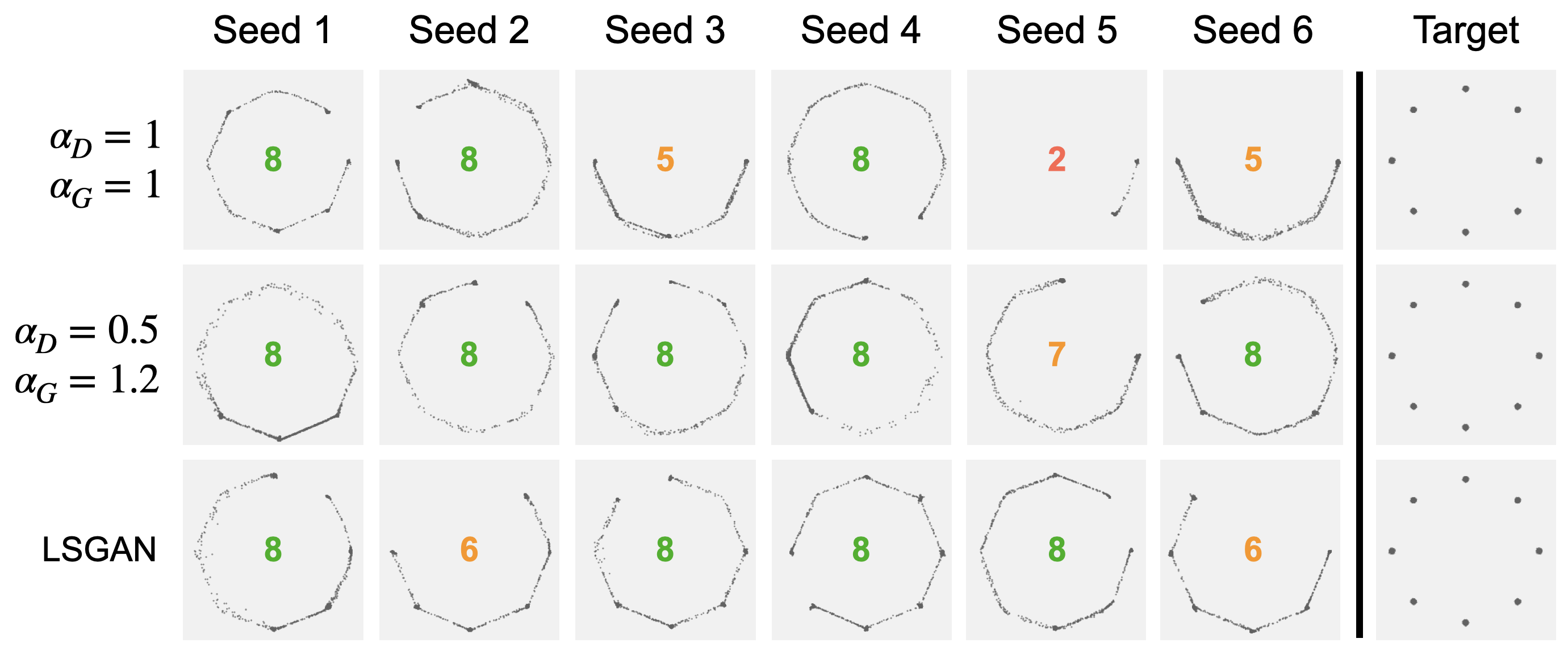}
    \caption{Generated samples from two $(\alpha_{D}, \alpha_{G})$-GANs trained with the \textbf{NS} objectives in \eqref{eqn:disc_obj}, \eqref{eqn:gen_obj_ns}, as well as the LSGAN. We provide 6 seeds to illustrate the stability in performance for each GAN across multiple runs.}
    \label{fig:NS}
    \vspace{-0.075in}
\end{figure}

For the saturating setting, the improvement in stability of the $(0.2,1)$-GAN relative to the vanilla GAN is illustrated in Fig. \ref{fig:sat-figure} as detailed in the caption. %Figure \ref{fig:sat-figure}(b) shows the significant gains achieved by using $\alpha_D<1$ in reducing instability. 
In fact, vanilla GAN completely fails to converge to the true distribution 30\% of the time while succeeding only 46\% of the time. In contrast, the $(\alpha_{D}, \alpha_{G})$-GAN with $\alpha_{D} < 1$ learns a more stable G due to a less confident D (see also Fig.~\ref{fig:sat-figure}(a)). For example, the $(0.3,1)$-GAN success and failure rates improve to 87\% and 2\%, respectively. 
Finally, for the NS setting in Fig. \ref{fig:NS}, we find that tuning $\alpha_D$ and $\alpha_G$ yields more consistently stable outcomes than vanilla and LSGANs. Mode coverage rates over 200 seeds for saturating  \if \extended 0%
    (Tables I and II) and NS (Table III) are in \cite[Appendix~E]{welfert2023towards}.
    \fi%
    \if \extended 1%
    (Tables \ref{table:2d-ring-sat-success-rates} and \ref{table:2d-ring-sat-failure-rates}) and NS (Table \ref{table:2d-ring-ns-success-rates}) are in Appendix \ref{appendix:experimental-details-results}.
    \fi

\vspace{-0.075in}
\subsection{Stacked MNIST}
\label{subsec:stacked-mnist}

The Stacked MNIST dataset is an enhancement of MNIST \cite{deng2012mnist} as it contains images of size $3 \times 28 \times 28$, where each RGB channel is a $28 \times 28$ image randomly sampled from MNIST. Stacked MNIST is a popular choice for image generation since its use of 3 channels allows for a total of $10^{3} = 1000$ modes, as opposed to the 10 modes (digits) in MNIST, which makes the latter much easier for GANs to learn. We generate 100,000 training samples, 25,000 testing samples, and the same number of 100-dimension latent Gaussian noise vectors.

We use the DCGAN architecture \cite{radford2015} for training, which uses deep convolutional neural networks (CNN) for both D and G (details in Tables \if \extended 0%
    IV, V\cite[Appendix~E]{welfert2023towards}).
    \fi%
    \if \extended 1%
    \ref{table:disc}, \ref{table:gen} of Appendix \ref{appendix:experimental-details-results}).
    \fi As in other works, we focus solely on the NS setting using appropriate objective functions for vanilla GAN, $(\alpha_{D}, \alpha_{G})$-GAN, and LSGAN. We compute the mode coverage of each trial by feeding each generated sample to a 1000-mode CNN classifier. %to classify as one of the 1000 modes. 
The classifier is obtained by pretraining on MNIST to achieve 99.5\% validation accuracy. We also consider a range of settings for two key hyperparameters: the number of epochs and learning rate for Adam optimization. %We consider a range of values for the number of epochs to train over; we also consider the following learning rates:  $10^{-4}$, $2 \times 10^{-4}$, $5 \times 10^{-4}$, $10^{-3}$, $2 \times 10^{-3}$, $5 \times 10^{-3}$, and $10^{-2}$ for Adam optimization. 
Each combination of objective function, number of epochs, and learning rate is trained for 100 seeds; this allows us to report the \emph{mean mode coverage}. We also report the mean Fréchet Inception Distance (FID)\footnote{FID is an unsupervised similarity metric between the real and generated feature distributions extracted by InceptionNet-V3~\cite{heusel2017fid}.}. %  over 100 seeds for each setting.

\begin{figure}[t]
    \centering
    \footnotesize
\setlength{\tabcolsep}{1pt}
\begin{tabular}{@{}cc@{}}
  \includegraphics[width=4.5cm]{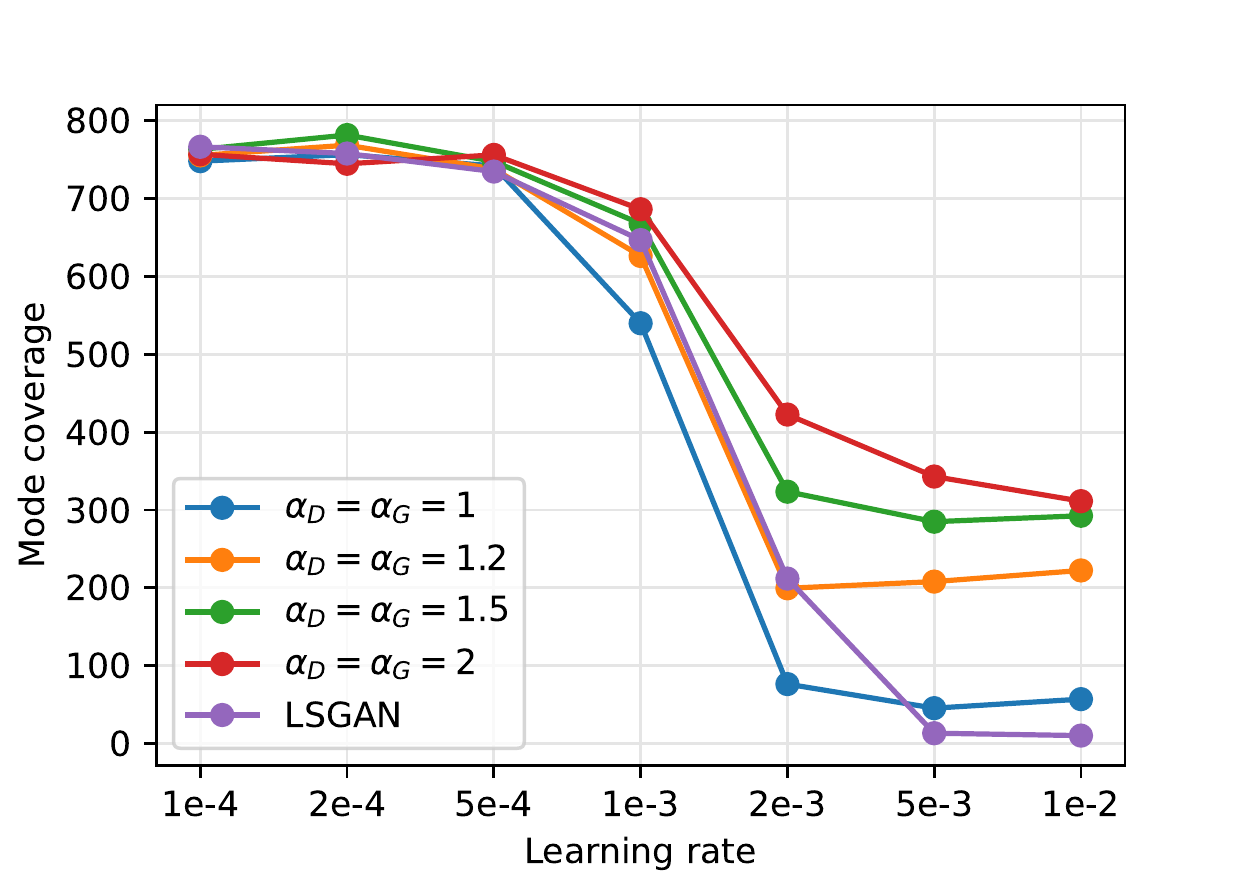}  & \includegraphics[width=4.5cm]{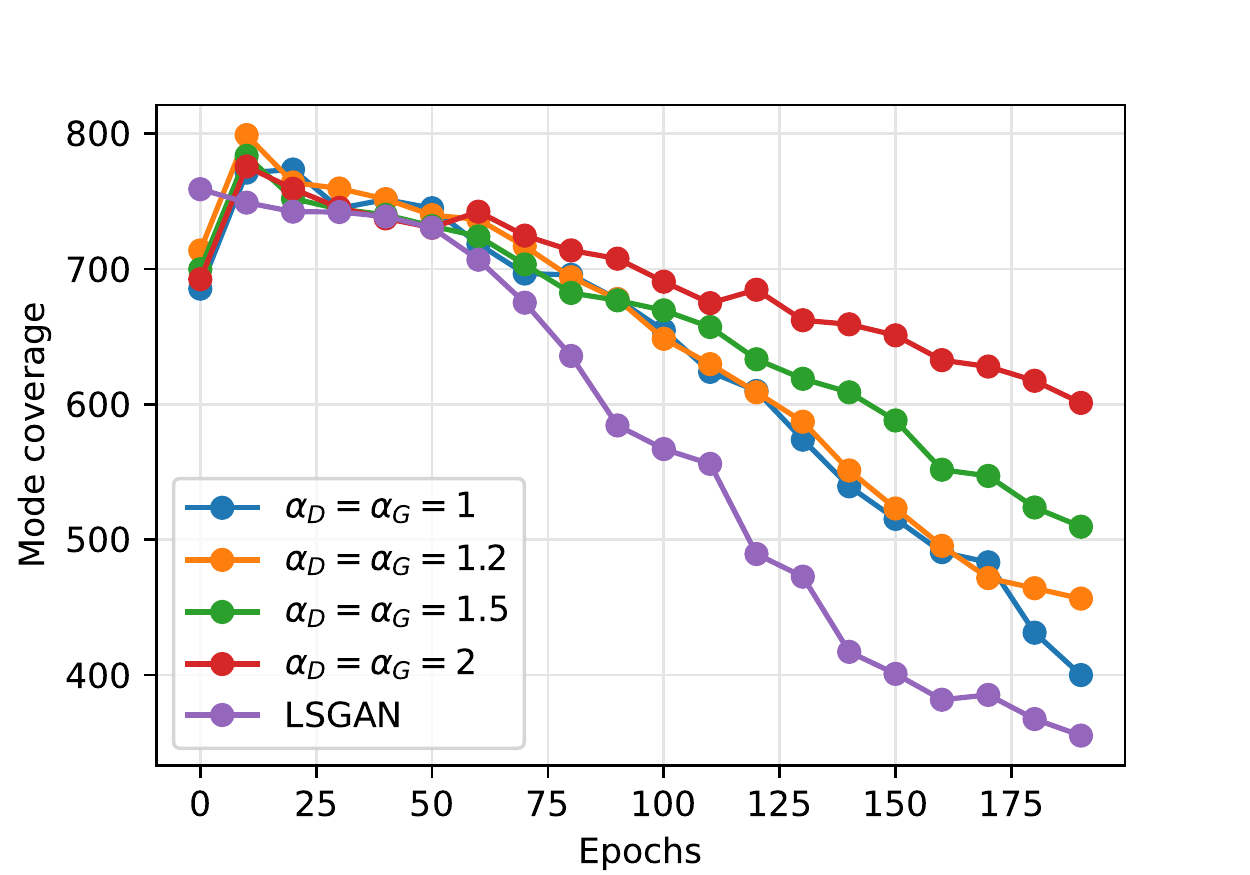} \\
   (a)  & (b)
\end{tabular}
\caption{Mode coverage vs. (a) varied learning rates with fixed epoch number ($=50$) and (b) varied epoch numbers with fixed learning rate ($=5\times 10^{-4}$) for different GANs, underscoring the vanilla GAN's hyperparameter sensitivity.
}
\label{fig:stacked-modes}
\end{figure}

\begin{figure}[t]
    \centering
    \includegraphics[width=8cm]{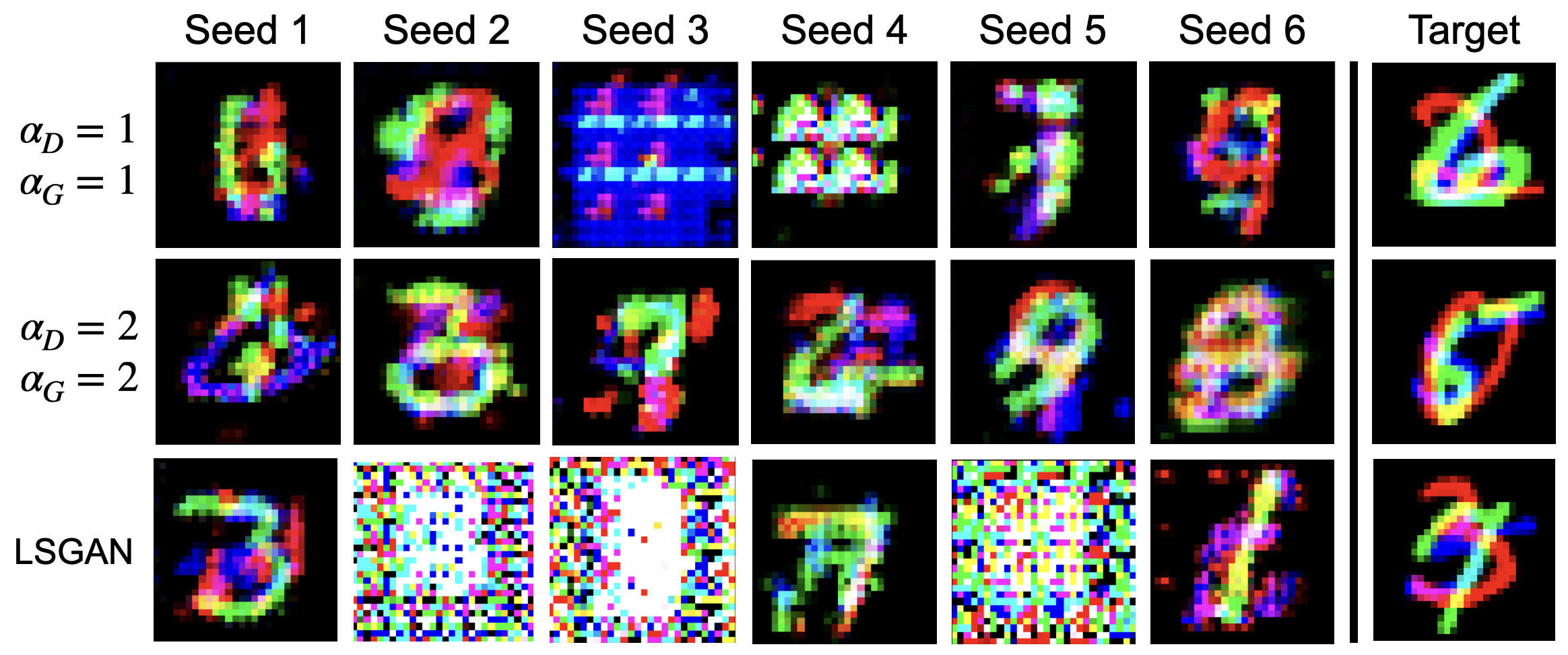}
    \caption{Generated Stacked MNIST samples from three GANs over 6 seeds when trained for 200 epochs with a learning rate of $5 \times 10^{-4}$.}
    \label{fig:stacked-output}
    \vspace{-0.075in}
\end{figure}

In Fig. \ref{fig:stacked-modes}(a) and \ref{fig:stacked-modes}(b), 
%With the Stacked MNIST dataset, 
we empirically demonstrate the dependence of mode coverage on learning rate and number of epochs, respectively (FID plots are in %Fig.~\ref{fig:stacked-fids} of 
\if \extended 0%
    \cite[Appendix~E-C]{welfert2023towards}).
    \fi%
    \if \extended 1%
    Appendix \ref{appendix:stacked-mnist}).
    \fi Achieving robustness to hyperparameter initialization is highly desirable in the unsupervised GAN setting as the choices that facilitate steady model convergence are not easily determined without prior mode knowledge. Observing the mode coverage of different $(\alpha_{D}, \alpha_{G})$-GANs, we find that as the learning rate or training time increases, the performance of both vanilla GAN and LSGAN deteriorates faster than a GAN with $\alpha_{D}= \alpha_{G} > 1$ (see \if \extended 0%
    \cite[Appendix~E]{welfert2023towards}
    \fi%
    \if \extended 1%
    Appendix \ref{appendix:experimental-details-results}
    \fi for additional details that motivate this choice). Finally, as shown in Fig. \ref{fig:stacked-output}, we observe that the outputs of $(\alpha_{D}, \alpha_{G})$-GAN are more consistent and accurate across multiple seeds, relative to LSGAN and vanilla GAN.

\section{Concluding Remarks}
We have introduced a dual-objective GAN formulation, focusing in particular on using $\alpha$-loss for both players' objectives. Our results highlight the value of tuning $\alpha$ in alleviating training instabilities and enhancing robustness to learning rates and training epochs, hyperparameters whose optimal values are generally not known \emph{a priori}. Generalization guarantees of $(\alpha_D,\alpha_G)$-GANs is a natural extension to study. An equally important problem is to evaluate if our observations hold more broadly, including, when the training data is noisy \cite{nietert2022outlier}.  

% \section{Acknowledgment}
% Gowtham R. Kurri was with Arizona State University when this work was done.
%The effect of $\alpha$ on gradients suggest that adaptively changing their values can further improve performance. 
\clearpage
\balance
\bibliographystyle{IEEEtran}
\bibliography{Bibliography}

\newpage
\if \extended 1
\appendices
\nobalance
\section{Proof of Theorem \ref{thm:alpha_D,alpha_G-GAN-saturating}}
\label{appendix:alpha_D,alpha_G-GAN-saturating}
The proof to obtain \eqref{eqn:optimaldisc-gen-alpha-GAN} is the same as that for \cite[Theorem~1]{KurriSS21}, where $\alpha=\alpha_D$. The generator's optimization problem in \eqref{eqn:gen_obj} with the optimal discriminator in \eqref{eqn:optimaldisc-gen-alpha-GAN} can be written as $\inf_{\theta\in\Theta}V_{\alpha_G}(\theta,\omega^*)$,
where
\begin{align*}
   &V_{\alpha_G}(\theta,\omega^*)\\&=\frac{\alpha_G}{\alpha_G-1} \times \\
   &\left[\int_\mathcal{X}\left(p_r(x)D_{\omega^*}(x)^{\frac{\alpha_G-1}{\alpha_G}}+p_{G_\theta}(x)(1-D_{\omega^*}(x))^{\frac{\alpha_G-1}{\alpha_G}}\right)dx-2\right]\\
   &=\frac{\alpha_G}{\alpha_G-1}\Bigg[\int_\mathcal{X}\Bigg(p_r(x)\left( \frac{p_r(x)^{\alpha_D}}{p_r(x)^{\alpha_D}+p_{G_\theta}(x)^{\alpha_D}}\right)^{\frac{\alpha_G-1}{\alpha_G}} \\
   & \qquad \qquad \quad+ p_{G_\theta}(x)\left( \frac{p_r(x)^{\alpha_D}}{p_r(x)^{\alpha_D}+p_{G_\theta}(x)^{\alpha_D}}\right)^{\frac{\alpha_G-1}{\alpha_G}}\Bigg)dx-2\Bigg]\\
    &=\frac{\alpha_G}{\alpha_G-1} \times \\
    &\left(\int_{\mathcal{X}}p_{G_\theta}(x)\left(\frac{(p_r(x)/p_{G_\theta}(x))^{\alpha_D(1-1/\alpha_G)+1}+1}{((p_r(x)/p_{G_\theta}(x))^{\alpha_D}+1)^{1-1/\alpha_G}}\right)dx-2\right) \\
    &  =   \int_\mathcal{X} p_{G_\theta}(x)f_{\alpha_D,\alpha_G}\left(\frac{p_r(x)}{p_{G_\theta}(x)}\right) dx + \frac{\alpha_G}{\alpha_G-1}\left(2^{\frac{1}{\alpha_G}}-2\right),
\end{align*}
where $f_{\alpha_D,\alpha_G}$ is as defined in \eqref{eqn:f-alpha_d,alpha_g}. Note that if $f_{\alpha_D,\alpha_G}$ is strictly convex, the first term in the last equality above equals an $f$-divergence which is minimized if and only if $P_r = P_{G_\theta}$. Define the regions $R_1$ and $R_2$ as follows:
\begin{align*}
    R_1 \coloneqq \Big\{(\alpha_D,\alpha_G) \in (0,\infty]^2 \bigm\vert \alpha_D \le 1,\alpha_G > \frac{\alpha_D}{\alpha_D+1}\Big\}
    % \label{eq:convex-reg1-sat}
\end{align*}
and 
\begin{align*}
    R_2 \coloneqq \Big\{(\alpha_D,\alpha_G) \in (0,\infty]^2 \bigm\vert \alpha_D > 1,\frac{\alpha_D}{2}< \alpha_G \le \alpha_D\Big\}.
    % \label{eq:convex-reg2-sat}
\end{align*}
In order to prove that $f_{\alpha_D,\alpha_G}$ is strictly convex for $(\alpha_D,\alpha_G)\in R_1\cup R_2$, we take its second derivative, which yields
\begin{align}
    &f^{\prime\prime}_{\alpha_D,\alpha_G}(u) \nonumber\\
    % & = \frac{\alpha_D}{\alpha_G}u^{\alpha_D-\frac{\alpha_D}{\alpha_G}-2}(1+u^{\alpha_D})^{\frac{1}{\alpha_G}-3}\bigg[\alpha_G(1+u^{\alpha_D}(u+u^\frac{\alpha_D}{\alpha_G}) \nonumber \\
    % & + \alpha_D\left((\alpha_G-1)u-\alpha_G u^{\alpha_D+1}+(\alpha_G-1)u^{\alpha_D+\frac{\alpha_D}{\alpha_G}}-\alpha_Gu^\frac{\alpha_D}{\alpha_G} \right)\bigg] \nonumber \\
    & = A_{\alpha_D,\alpha_G}(u) \bigg[(\alpha_G+\alpha_D\alpha_G-\alpha_D)\left(u+u^{\alpha_D+\frac{\alpha_D}{\alpha_G}}\right) \nonumber \\
    & \qquad \qquad \qquad \qquad+(\alpha_G-\alpha_D\alpha_G)\left(u^\frac{\alpha_D}{\alpha_G}+u^{\alpha_D+1}\right)\bigg],
    \label{eq:sec_deriv_sat_sym}
\end{align}
where 
\begin{align}
    &A_{\alpha_D,\alpha_G}(u)=\frac{\alpha_D}{\alpha_G}u^{\alpha_D-\frac{\alpha_D}{\alpha_G}-2}(1+u^{\alpha_D})^{\frac{1}{\alpha_G}-3}.
    \label{eq:f-sec-deriv-mult-const}
\end{align}
Note that $A_{\alpha_D,\alpha_G}(u)> 0$ for all $u > 0$ and $\alpha_D,\alpha_G\in(0,\infty]$. Therefore, in order to ensure $f^{\prime\prime}_{\alpha_D,\alpha_G}(u)>0$ for all $u>0$ it is sufficient to have
\begin{align}
    \alpha_G+\alpha_D\alpha_G-\alpha_D > \alpha_G(\alpha_D-1)B_{\alpha_D,\alpha_G}(u),
    \label{eq:sat-main-convexity-condition}
\end{align}
where
\begin{align}
    B_{\alpha_D,\alpha_G}(u) = \frac{u^\frac{\alpha_D}{\alpha_G}+u^{\alpha_D+1}}{u+u^{\alpha_D+\frac{\alpha_D}{\alpha_G}}}
    \label{eq:B-function}
\end{align} for $u > 0$. Since $B_{\alpha_D,\alpha_G}(u) > 0$ for all $u  > 0$, the sign of the RHS of \eqref{eq:sat-main-convexity-condition} is determined by whether $\alpha_D \le 1$ or $\alpha_D > 1$. We look further into these two cases in the following:

\noindent\textbf{Case 1:} $\alpha_D \le 1$. Then $\alpha_G(\alpha_D-1)B_{\alpha_D,\alpha_G}(u) \le 0$ for all $u > 0$ and $(\alpha_D,\alpha_G)\in(0,\infty]^2$. Therefore, we need
\begin{align}
    \alpha_G(1+\alpha_D)-\alpha_D > 0 \Leftrightarrow \alpha_G > \frac{\alpha_D}{\alpha_D+1}.
\end{align}
\noindent\textbf{Case 2:} $\alpha_D > 1$. Then $\alpha_G(\alpha_D-1)B_{\alpha_D,\alpha_G}(u) > 0$ for all $u > 0$ and $(\alpha_D,\alpha_G)\in(0,\infty]^2$. In order to obtain conditions on $\alpha_D$ and $\alpha_G$, we determine the monotonicity of $B_{\alpha_D,\alpha_G}$ by finding its first derivative as follows:
% \begin{align}
%     &B^\prime_{\alpha_D,\alpha_G}(u) \\
%     &= \frac{(\alpha_G-\alpha_D)\Big(u^{\frac{\alpha_D}{\alpha_G}+2\alpha_D}-u^{\frac{\alpha_D}{\alpha_G}}\Big)+\alpha_D\alpha_G\Big(u^{\alpha_D+1}-u^{2\frac{\alpha_D}{\alpha_G}+\alpha_D-1}\Big)}{\alpha_G\Big(u+u^{\frac{\alpha_D}{\alpha_G}+\alpha_D}\Big)^2}
% \end{align}
\begin{align*}
    &B^\prime_{\alpha_D,\alpha_G}(u) \nonumber \\
    &= \frac{(\alpha_G-\alpha_D)(u^{2\alpha_D}-1)+\alpha_D\alpha_G\Big(u^{\alpha_D-\frac{\alpha_D}{\alpha_G}+1}-u^{\alpha_D+\frac{\alpha_D}{\alpha_G}-1}\Big)}{\alpha_G u^{-\frac{\alpha_D}{\alpha_G}}\Big(u+u^{\alpha_D+\frac{\alpha_D}{\alpha_G}}\Big)^2}.
    \label{eq:B-deriv}
\end{align*}
Since the denominator of $B^\prime_{\alpha_D,\alpha_G}$ is positive for all $u>0$ and $(\alpha_D,\alpha_G)\in (0,\infty]^2$, we just need to check the sign of the numerator.
\\\textbf{Case 2a:} $\alpha_D>\alpha_G$. For $u \in (0,1)$, 
\[u^{2\alpha_D}-1 < 0 \quad \text{and} 
 \quad u^{\alpha_D-\frac{\alpha_D}{\alpha_G}+1}-u^{\alpha_D+\frac{\alpha_D}{\alpha_G}-1} >0,\]
so $B^\prime_{\alpha_D,\alpha_G}(u) > 0$. For $u > 1$, 
\[u^{2\alpha_D}-1 > 0 \quad \text{and} 
 \quad u^{\alpha_D-\frac{\alpha_D}{\alpha_G}+1}-u^{\alpha_D+\frac{\alpha_D}{\alpha_G}-1} < 0,\]
 so $B^\prime_{\alpha_D,\alpha_G}(u) < 0$. For $u=1$, $B^\prime_{\alpha_D,\alpha_G}(u) = 0$.  Hence, $B^\prime_{\alpha_D,\alpha_G}$ is strictly increasing for $u\in (0,1)$ and strictly decreasing for $u \ge 1$. Therefore, $B_{\alpha_D,\alpha_G}$ attains a maximum value of 1 at $u=1$. This means $B_{\alpha_D,\alpha_G}$ is bounded, i.e. $B_{\alpha_D,\alpha_G}\in (0,1]$ for all $u>0$. Thus, in order for \eqref{eq:sat-main-convexity-condition} to hold, it suffices to ensure that
\begin{align}
    \alpha_G+\alpha_D\alpha_G-\alpha_D > \alpha_G(\alpha_D-1) \Leftrightarrow \alpha_G > \frac{\alpha_G}{2}.
\end{align}
\textbf{Case 2b:} $\alpha_D<\alpha_G$. For $u \in (0,1)$, $u^{2\alpha_D}-1 < 0$ and $u^{\alpha_D-\frac{\alpha_D}{\alpha_G}+1}-u^{\alpha_D+\frac{\alpha_D}{\alpha_G}-1} < 0$, so $B^\prime_{\alpha_D,\alpha_G}(u) < 0$. For $u > 1$, $u^{2\alpha_D}-1 > 0$ and $u^{\alpha_D-\frac{\alpha_D}{\alpha_G}+1}-u^{\alpha_D+\frac{\alpha_D}{\alpha_G}-1} > 0$, so $B^\prime_{\alpha_D,\alpha_G}(u) > 0$. Hence, $B^\prime_{\alpha_D,\alpha_G}$ is strictly decreasing for $u\in (0,1)$ and strictly increasing for $u \ge 1$. Therefore, $B_{\alpha_D,\alpha_G}$ attains a minimum value of 1 at $u=1$. This means that $B_{\alpha_D,\alpha_G}$ is not bounded above, so it is not possible to satisfy \eqref{eq:sat-main-convexity-condition} without restricting the domain of $B_{\alpha_D,\alpha_G}$.

% and
% \[\alpha_G-\alpha_D\alpha_G \ge 0 \Leftrightarrow \alpha_D \le 1. \]
% \begin{align}
%     &f^{\prime\prime}_{\alpha_D,\alpha_G}(u) \nonumber\\
%     & = \frac{\alpha_D}{\alpha_G}u^{\alpha_D-\frac{\alpha_D}{\alpha_G}-2}(1+u^{\alpha_D})^{\frac{1}{\alpha_G}-3}\bigg[\alpha_G(1+u^{\alpha_D}(u+u^\frac{\alpha_D}{\alpha_G}) \nonumber \\
%     & + \alpha_D\left((\alpha_G-1)u-\alpha_G u^{\alpha_D+1}+(\alpha_G-1)u^{\alpha_D+\frac{\alpha_D}{\alpha_G}}-\alpha_Gu^\frac{\alpha_D}{\alpha_G} \right)\bigg] \nonumber \\
%     & = A_{\alpha_D,\alpha_G}(u) \bigg[(\alpha_G+\alpha_D\alpha_G-\alpha_D)B_{\alpha_D,\alpha_G}(u) \nonumber \\
%     & \qquad+(\alpha_G-\alpha_D\alpha_G)C_{\alpha_D,\alpha_G}(u)\bigg].
%     \label{eq:sec_deriv_sat_sym}
% \end{align}
% where 
% \begin{align*}
%     &A_{\alpha_D,\alpha_G}(u)=\frac{\alpha_D}{\alpha_G}u^{\alpha_D-\frac{\alpha_D}{\alpha_G}-2}(1+u^{\alpha_D})^{\frac{1}{\alpha_G}-3}, \\
%     & B_{\alpha_D,\alpha_G}(u)=u+u^{\alpha_D+\frac{\alpha_D}{\alpha_G}}, \\
%     & C_{\alpha_D,\alpha_G}(u) = u^\frac{\alpha_D}{\alpha_G}+u^{\alpha_D+1}.
% \end{align*}
% Note that $A_{\alpha_D,\alpha_G}(u)> 0$, $B_{\alpha_D,\alpha_G}(u)> 0$, and $C_{\alpha_D,\alpha_G}(u)> 0$ for all $u > 0$, $\alpha_D,\alpha_G\in(0,\infty]$. Therefore, in order to ensure $f^{\prime\prime}_{\alpha_D,\alpha_G}(u)>0$ for all $u>0$ it is sufficient to have
% \[\alpha_G+\alpha_D\alpha_G-\alpha_D > 0 \Leftrightarrow \alpha_G > \frac{\alpha_G}{\alpha_G+1} \]
% and
% \[\alpha_G-\alpha_D\alpha_G \ge 0 \Leftrightarrow \alpha_D \le 1. \]
 Thus, for $(\alpha_D,\alpha_G)\in R_1 \cup R_2$,
\[
   V_{\alpha_G}(\theta,\omega^*)
    =D_{f_{\alpha_D,\alpha_G}}(P_r||P_{G_\theta})+\frac{\alpha_G}{\alpha_G-1}\left(2^{\frac{1}{\alpha_G}}-2\right).
\]
This yields \eqref{eqn:gen-alpha_d,alpha_g-obj}. Note that $D_{f_{\alpha_D,\alpha_G}}(P||Q)$ is symmetric since
% \begin{align*}
%     &f_{\alpha_D,\alpha_G}(1/u) \\
%     &= \frac{\alpha_G}{\alpha_G-1}\left(\frac{u^{-\alpha_d(1-1/\alpha_G)-1}+1}{(u^{-\alpha_D}+1)^{1-1/\alpha_G}}-2^{\frac{1}{\alpha_G}}\right)\\
%     &=\frac{\alpha_G}{\alpha_G-1}\left(\frac{u^{-\alpha_d(1-1/\alpha_G)-1}+1}{u^{-\alpha_d(1-1/\alpha_G)}(1+u^{\alpha_D})^{1-1/\alpha_G}}-2^{\frac{1}{\alpha_G}}\right) \\
%     & = \frac{\alpha_G}{\alpha_G-1}\left(\frac{u^{-1}+u^{\alpha_d(1-1/\alpha_G)}}{(1+u^{\alpha_D})^{1-1/\alpha_G}}-2^{\frac{1}{\alpha_G}}\right) 
% \end{align*}
\begin{align*}
    &D_{f_{\alpha_D,\alpha_G}}(Q||P) \\
    &= \int_\mathcal{X} p(x)f_{\alpha_D,\alpha_G}\left(\frac{q(x)}{p(x)}\right) dx \\
    &=\frac{\alpha_G}{\alpha_G-1} \times \\
    &\left(\int_{\mathcal{X}}p(x)\left(\frac{(p(x)/q(x))^{-{\alpha_D\left(1-\frac{1}{\alpha_G}\right)}-1}+1}{((p(x)/q(x))^{-\alpha_D}+1)^{1-\frac{1}{\alpha_G}}}\right)dx-2^\frac{1}{\alpha_G}\right) \\
    &=\frac{\alpha_G}{\alpha_G-1} \times \\
    &\left(\int_{\mathcal{X}}p(x)\left(\frac{q(x)/p(x)+(p(x)/q(x))^{\alpha_D\left(1-\frac{1}{\alpha_G}\right)}}{(1+(p(x)/q(x))^{\alpha_D})^{1-\frac{1}{\alpha_G}}}\right)dx-2^\frac{1}{\alpha_G}\right)\\
    &=\frac{\alpha_G}{\alpha_G-1} \times \\
    &\left(\int_{\mathcal{X}}q(x)\left(\frac{1+(p(x)/q(x))^{\alpha_D\left(1-\frac{1}{\alpha_G}\right)}}{(1+(p(x)/q(x))^{\alpha_D})^{1-\frac{1}{\alpha_G}}}\right)dx-2^\frac{1}{\alpha_G}\right)\\
    & = D_{f_{\alpha_D,\alpha_G}}(P||Q).
\end{align*}
Since $f_{\alpha_D,\alpha_G}$ is strictly convex and $f_{\alpha_D,\alpha_G}(1)=0$, $D_{f_{\alpha_D,\alpha_G}}(P_r||P_{G_\theta})\geq 0$ with equality if and only if $P_r=P_{G_\theta}$. Thus, we have $V_{\alpha_G}(\theta,\omega^*)\geq \frac{\alpha_G}{\alpha_G-1}\left(2^{\frac{1}{\alpha_G}}-2\right)$ with equality if and only if $P_r=P_{G_\theta}$.

%     \begin{figure}[ht]
% \centering
% \begin{tikzpicture}
%     \begin{axis}[axis lines = middle, 
%     xlabel = $\alpha_D$, ylabel = $\alpha_G$,
%     x label style={at={(axis description cs:1.07,0.13)},anchor=west},
%     y label style={at={(axis description cs:0.14,1.07)},anchor=south},
% 	xmin = -0.2, xmax = 2.7, ymin = -0.2, ymax = 2.7,
% 	xtick={1,2},
%     xticklabels={1,2},
%     ytick={1,2},
%     yticklabels={1,2},
% 	anchor=origin,  
% 	x=2cm, y=2cm, 
% 	]
% 		\addplot[name path=b, domain = 0:1, samples = 200, smooth, draw=none] {2.5};
% 		\addplot[name path=a, domain = 0:1, samples = 200, dashed, ultra thick, blue] {x/(x+1)}
% 		node [right,pos=1] {\large $\textcolor{blue}{\alpha_G=\frac{\alpha_D}{\alpha_D+1}}$};
% 		\addplot +[mark=none,smooth,red,ultra thick] coordinates {(1, 0) (1, 2.5)};
% 		\addplot[name path=c, domain = 1:2.5, samples = 200, smooth, ultra thick, blue] {x};
% 		\addplot[name path=d, domain = 1:2.5, samples = 200, dashed, ultra thick, blue] {x/2};
% 	\addplot [
%     			thick,
%     			color=green,
%     			fill=cyan, 
%     			fill opacity=0.4
% 		]
% 		fill between[of=b and a, soft clip={domain=0:5}];
% 	\addplot [
%     			thick,
%     			color=green,
%     			fill=cyan, 
%     			fill opacity=0.4
% 		]
% 		fill between[of=c and d, soft clip={domain=0:5}];
% 	\node [xshift=1cm, yshift=2.7cm] {\large $R$};
% \end{axis}
% \end{tikzpicture}
% \caption{Plot of region $R=\{(\alpha_D,\alpha_G)\in (0,\infty]^2 \mid \alpha_D \le 1, \alpha_G > \frac{\alpha_D}{\alpha_D+1}\}$ for which $f_{\alpha_D,\alpha_G}$ is strictly convex.}
% \label{fig:convexity_region_sat_sym}
% \end{figure}

\section{Proof of Theorem \ref{thm:alpha_D,alpha_G-GAN-nonsaturating}}
\label{appendix:alpha_D,alpha_G-GAN-nonsaturating}
The generator's optimization problem in \eqref{eqn:gen_obj} with the optimal discriminator in \eqref{eqn:optimaldisc-gen-alpha-GAN} can be written as $\inf_{\theta\in\Theta}V^\text{NS}_{\alpha_G}(\theta,\omega^*)$,
where
\begin{align*}
   &V^\text{NS}_{\alpha_G}(\theta,\omega^*) \\
   &=\frac{\alpha_G}{\alpha_G-1}\left[1-\int_\mathcal{X}\left(p_{G_\theta}(x)D_{\omega^*}(x)^{\frac{\alpha_G-1}{\alpha_G}}\right)dx\right]\\
   &=\frac{\alpha_G}{\alpha_G-1}\Bigg[1-\int_\mathcal{X}p_{G_\theta}(x)\left( \frac{p_r(x)^{\alpha_D}}{p_r(x)^{\alpha_D}+p_{G_\theta}(x)^{\alpha_D}}\right)^{\frac{\alpha_G-1}{\alpha_G}}dx\Bigg]\\
    &=\frac{\alpha_G}{\alpha_G-1}\Bigg[1-\int_\mathcal{X}p_{G_\theta}(x) \frac{(p_r(x)/p_{G_\theta}(x))^{\alpha_D(1-1/\alpha_G)}}{((p_r(x)/p_{G_\theta}(x))^{\alpha_D}+1)^{1-1/\alpha_G}}dx\Bigg] \\
    & = \int_\mathcal{X} p_{G_\theta}(x)f^\text{NS}_{\alpha_D,\alpha_G}\left(\frac{p_r(x)}{p_{G_\theta}(x)}\right) dx + \frac{\alpha_G}{\alpha_G-1}\left(1-2^{\frac{1}{\alpha_G}-1}\right),
\end{align*}
where $f^\text{NS}_{\alpha_D,\alpha_G}$ is as defined in \eqref{eqn:f-alpha_d,alpha_g-ns}. In order to prove that $f^\text{NS}_{\alpha_D,\alpha_G}$ is strictly convex for $(\alpha_D,\alpha_G)\in R_\text{NS}= \{(\alpha_D,\alpha_G) \in (0,\infty]^2 \mid \alpha_D > \alpha_G(\alpha_D-1)\}$, we take its second derivative, which yields
\begin{align}
    &f^{\prime\prime}_{\alpha_D,\alpha_G}(u) \nonumber\\
    & = A_{\alpha_D,\alpha_G}(u) \bigg[(\alpha_G-\alpha_D\alpha_G+\alpha_D) +\alpha_G(1+\alpha_D)u^{\alpha_D}\bigg],
    \label{eq:sec_deriv_nonsat}
\end{align}
where $A_{\alpha_D,\alpha_G}$ is defined as in \eqref{eq:f-sec-deriv-mult-const}.
Since $A_{\alpha_D,\alpha_G}(u)> 0$ for all $u > 0$ and $(\alpha_D,\alpha_G)\in(0,\infty]^2$, to ensure $f^{\prime\prime}_{\alpha_D,\alpha_G}(u)>0$ for all $u>0$ it suffices to have
\[\frac{\alpha_G-\alpha_D\alpha_G+\alpha_D}{\alpha_G(1+\alpha_D)} > -u^{\alpha_D} \]
for all $u > 0$. This is equivalent to
\[\frac{\alpha_G-\alpha_D\alpha_G+\alpha_D}{\alpha_G(1+\alpha_D)} > 0, \]
which results in the condition
\[ \alpha_D > \alpha_G(\alpha_D-1)\]
for $(\alpha_D,\alpha_G)\in(0,\infty]^2$. Thus, for $(\alpha_D,\alpha_G)\in R_\text{NS}$,
\[
   V^\text{NS}_{\alpha_G}(\theta,\omega^*)
    =D_{f^\text{NS}_{\alpha_D,\alpha_G}}(P_r||P_{G_\theta})+\frac{\alpha_G}{\alpha_G-1}\left(1-2^{\frac{1}{\alpha_G}-1}\right).
\]
This yields \eqref{eqn:gen-alpha_d,alpha_g-obj-ns}. Note that $D_{f^\text{NS}_{\alpha_D,\alpha_G}}(P||Q)$ is not symmetric since $D_{f^\text{NS}_{\alpha_D,\alpha_G}}(P||Q) \ne D_{f^\text{NS}_{\alpha_D,\alpha_G}}(Q||P)$.
Since $f^\text{NS}_{\alpha_D,\alpha_G}$ is strictly convex and $f^\text{NS}_{\alpha_D,\alpha_G}(1)=0$, $D_{f^\text{NS}_{\alpha_D,\alpha_G}}(P_r||P_{G_\theta})\geq 0$ with equality if and only if $P_r=P_{G_\theta}$. Thus, we have $V^\text{NS}_{\alpha_G}(\theta,\omega^*)\geq \frac{\alpha_G}{\alpha_G-1}\left(1-2^{\frac{1}{\alpha_G}-1}\right)$ with equality if and only if $P_r=P_{G_\theta}$. 

\section{Proof of Theorem \ref{thm:estimationerror-upperbound-alpha_d,alpha_g-GAN}}
\label{appendix:estimationerror-upperbound-alpha_d,alpha_g-GAN}
By adding and subtracting relevant terms, we obtain
% \begin{align}
%     &d_{\omega^*(\hat{\theta}^*)}(P_r,{P}_{G_{\hat{\theta}^*}})-\inf_{\theta\in\Theta} d_{\omega^*(\theta)}(P_r,P_{G_{\theta}}) \nonumber \\
%     & = \underbrace{d_{\omega^*(\hat{\theta}^*)}(P_r,{P}_{G_{\hat{\theta}^*}}) - d_{\omega^*(\hat{\theta}^*)}(\hat{P}_r,{P}_{G_{\hat{\theta}^*}})}_{\text{(I)}} \\
%     & \quad + \underbrace{\inf_{\theta\in\Theta} d_{\omega^*(\theta)}(\hat{P}_r,P_{G_{\theta}}) - \inf_{\theta\in\Theta} d_{\omega^*(\theta)}(P_r,P_{G_{\theta}})}_{\text{(II)}} \\
%     & \quad + \underbrace{d_{\omega^*(\hat{\theta}^*)}(\hat{P}_r,{P}_{G_{\hat{\theta}^*}}) - \inf_{\theta\in\Theta} d_{\omega^*(\theta)}(\hat{P}_r,P_{G_{\theta}})}_{\text{(III)}}.
%     \label{eq:estimation-err-expanded}
% \end{align}
\begin{subequations}
\begin{align} 
&d_{\omega^*(P_r,P_{G_{\hat{\theta}^*}})}(P_r,{P}_{G_{\hat{\theta}^*}})-\inf_{\theta\in\Theta} d_{\omega^*(P_r,P_{G_{\theta}})}(P_r,P_{G_{\theta}}) \nonumber \\
& = d_{\omega^*(P_r,P_{G_{\hat{\theta}^*}})}(P_r,{P}_{G_{\hat{\theta}^*}}) - d_{\omega^*(P_r,P_{G_{\hat{\theta}^*}})}(\hat{P}_r,{P}_{G_{\hat{\theta}^*}}) \label{eq:estimation-err-expanded-term1} \\
&\quad + \inf_{\theta\in\Theta} d_{\omega^*(P_r,P_{G_{\theta}})}(\hat{P}_r,P_{G_{\theta}}) - \inf_{\theta\in\Theta} d_{\omega^*(P_r,P_{G_{\theta}})}(P_r,P_{G_{\theta}}) \label{eq:estimation-err-expanded-term2} \\
& \quad + d_{\omega^*(P_r,P_{G_{\hat{\theta}^*}})}(\hat{P}_r,{P}_{G_{\hat{\theta}^*}}) - \inf_{\theta\in\Theta} d_{\omega^*(P_r,P_{G_{\theta}})}(\hat{P}_r,P_{G_{\theta}}). \label{eq:estimation-err-expanded-term3}
\end{align}
\label{eq:estimation-err-expanded}
\end{subequations}

We upper-bound \eqref{eq:estimation-err-expanded} in the following three steps. Let $\phi(\cdot) = -\ell_{\alpha_G}(1,\cdot)$ and $\psi(\cdot) = -\ell_{\alpha_G}(0,\cdot)$.

We first upper-bound \eqref{eq:estimation-err-expanded-term1}. Let $\omega^*(\hat{\theta}^*)=\omega^*(P_r,P_{G_{\hat{\theta}^*}})$. Using \eqref{eq:est-err-gen-obj} yields
\begin{align}
    & d_{\omega^*(P_r,P_{G_{\hat{\theta}^*}})}(P_r,{P}_{G_{\hat{\theta}^*}}) - d_{\omega^*(P_r,P_{G_{\hat{\theta}^*}})}(\hat{P}_r,{P}_{G_{\hat{\theta}^*}}) \nonumber \\
    & = \mathbb{E}_{X\sim P_r}[\phi(D_{\omega^*(\hat{\theta}^*)}(X))] +\mathbb{E}_{X\sim P_{G_{\hat{\theta}^*}}}[\psi(D_{\omega^*(\hat{\theta}^*)}(X))] \nonumber \\
    & \quad - \left(\mathbb{E}_{X\sim \hat{P}_r}[\phi(D_{\omega^*(\hat{\theta}^*)}(X))] + \mathbb{E}_{X\sim P_{G_{\hat{\theta}^*}}}[\psi(D_{\omega^*(\hat{\theta}^*)}(X))] \right) \nonumber \\
    % & = \mathbb{E}_{X\sim P_r}[\phi(D_{\omega^*(\hat{\theta}^*)}(X))] - \mathbb{E}_{X\sim \hat{P}_r}[\phi(D_{\omega^*(\hat{\theta}^*)}(X))] \nonumber \\
    & \le \left| \mathbb{E}_{X\sim P_r}[\phi(D_{\omega^*(\hat{\theta}^*)}(X))] - \mathbb{E}_{X\sim \hat{P}_r}[\phi(D_{\omega^*(\hat{\theta}^*)}(X))] \right| \nonumber \\
    & \le \sup_{\omega \in \Omega} \left| \mathbb{E}_{X\sim P_r}[\phi(D_{\omega}(X))] - \mathbb{E}_{X\sim \hat{P}_r}[\phi(D_{\omega}(X))] \right|.
    \label{eq:est-err-term1-bound}
\end{align}

Next, we upper-bound \eqref{eq:estimation-err-expanded-term2}. Let $\theta^* = \arg\min_{\theta \in \Theta} d_{\omega^*(P_r,P_{G_{{\theta}}})}(P_r,{P}_{G_{{\theta}}})$ and $\omega^*({\theta}^*)=\omega^*(P_r,P_{G_{{\theta}^*}})$. Then
\begin{align}
    & \inf_{\theta\in\Theta} d_{\omega^*(P_r,P_{G_{{\theta}}})}(\hat{P}_r,P_{G_{\theta}}) - \inf_{\theta\in\Theta} d_{\omega^*(P_r,P_{G_{{\theta}}})}(P_r,P_{G_{\theta}}) \nonumber \\
    & \le d_{\omega^*(\theta^*)}(\hat{P}_r,P_{G_{\theta^*}}) - d_{\omega^*(\theta^*)}(P_r,P_{G_{\theta^*}}) \nonumber \\
    & = \mathbb{E}_{X\sim \hat{P}_r}[\phi(D_{\omega^*({\theta}^*)}(X))] +\mathbb{E}_{X\sim P_{G_{{\theta}^*}}}[\psi(D_{\omega^*({\theta}^*)}(X))] \nonumber \\
    & \quad - \left(\mathbb{E}_{X\sim {P}_r}[\phi(D_{\omega^*({\theta}^*)}(X))] + \mathbb{E}_{X\sim P_{G_{{\theta}^*}}}[\psi(D_{\omega^*({\theta}^*)}(X))] \right) \nonumber \\
    & = \mathbb{E}_{X\sim \hat{P}_r}[\phi(D_{\omega^*({\theta}^*)}(X))] - \mathbb{E}_{X\sim {P}_r}[\phi(D_{\omega^*({\theta}^*)}(X))] \nonumber \\
    & \le \sup_{\omega \in \Omega} \left| \mathbb{E}_{X\sim P_r}[\phi(D_{\omega}(X))] - \mathbb{E}_{X\sim \hat{P}_r}[\phi(D_{\omega}(X))] \right|.
    \label{eq:est-err-term2-bound}
\end{align}

Lastly, we upper-bound \eqref{eq:estimation-err-expanded-term3}. Let $\Tilde{\theta} = \arg\min_{\theta \in \Theta} d_{\omega^*(P_r,P_{G_{\theta}})}(\hat{P}_r,{P}_{G_{{\theta}}})$ and $\omega^*(\Tilde{\theta})=\omega^*(P_r,P_{G_{\Tilde{\theta}}})$. Then
\begin{align}
    & d_{\omega^*(P_r,P_{G_{\hat{\theta}^*}})}(\hat{P}_r,{P}_{G_{\hat{\theta}^*}}) - \inf_{\theta\in\Theta} d_{\omega^*(P_r,P_{G_{\theta}})}(\hat{P}_r,P_{G_{\theta}}) \nonumber \\
    & = d_{\omega^*(\hat{\theta}^*)}(\hat{P}_r,{P}_{G_{\hat{\theta}^*}}) - d_{\omega^*({\Tilde{\theta}})}(\hat{P}_r,\hat{P}_{G_{\Tilde{\theta}}}) \nonumber \\
    & \quad + d_{\omega^*({\Tilde{\theta}})}(\hat{P}_r,\hat{P}_{G_{\Tilde{\theta}}}) - d_{\omega^*(\Tilde{\theta})}(\hat{P}_r,P_{G_{\Tilde{\theta}}}) \nonumber \\
    & \le d_{\omega^*(\hat{\theta}^*)}(\hat{P}_r,{P}_{G_{\hat{\theta}^*}}) - d_{\omega^*({\hat{\theta}^*})}(\hat{P}_r,\hat{P}_{G_{\hat{\theta}^*}}) \nonumber \\
    & \quad + d_{\omega^*({\Tilde{\theta}})}(\hat{P}_r,\hat{P}_{G_{\Tilde{\theta}}}) - d_{\omega^*(\Tilde{\theta})}(\hat{P}_r,P_{G_{\Tilde{\theta}}}) \nonumber \\
    & = \mathbb{E}_{X\sim \hat{P}_r}[\phi(D_{\omega^*(\hat{\theta}^*)}(X))] +\mathbb{E}_{X\sim P_{G_{\hat{\theta}^*}}}[\psi(D_{\omega^*(\hat{\theta}^*)}(X))] \nonumber \\
    & \quad - \left(\mathbb{E}_{X\sim \hat{P}_r}[\phi(D_{\omega^*(\hat{\theta}^*)}(X))] + \mathbb{E}_{X\sim \hat{P}_{G_{\hat{\theta}^*}}}[\psi(D_{\omega^*(\hat{\theta}^*)}(X))] \right) \nonumber \\
    & \quad + \mathbb{E}_{X\sim \hat{P}_r}[\phi(D_{\omega^*(\Tilde{\theta})}(X))] +\mathbb{E}_{X\sim \hat{P}_{G_{\Tilde{\theta}}}}[\psi(D_{\omega^*(\Tilde{\theta})}(X))] \nonumber \\
    & \quad - \left(\mathbb{E}_{X\sim \hat{P}_r}[\phi(D_{\omega^*(\Tilde{\theta})}(X))] + \mathbb{E}_{X\sim {P}_{G_{\Tilde{\theta}}}}[\psi(D_{\omega^*(\Tilde{\theta})}(X))] \right) \nonumber \\
    & = \mathbb{E}_{X\sim P_{G_{\hat{\theta}^*}}}[\psi(D_{\omega^*(\hat{\theta}^*)}(X))] - \mathbb{E}_{X\sim \hat{P}_{G_{\hat{\theta}^*}}}[\psi(D_{\omega^*(\hat{\theta}^*)}(X))] \nonumber \\
    & \quad + \mathbb{E}_{X\sim \hat{P}_{G_{\Tilde{\theta}}}}[\psi(D_{\omega^*(\Tilde{\theta})}(X))] - \mathbb{E}_{X\sim {P}_{G_{\Tilde{\theta}}}}[\psi(D_{\omega^*(\Tilde{\theta})}(X))] \nonumber \\
    & \le 2\sup_{\omega \in \Omega,\theta \in \Theta} \left| \mathbb{E}_{X\sim P_{G_{{\theta}}}}[\psi(D_{\omega}(X))] - \mathbb{E}_{X\sim \hat{P}_{G_{{\theta}}}}[\psi(D_{\omega}(X))] \right|.
    \label{eq:est-err-term3-bound}
\end{align}
Combining \eqref{eq:est-err-term1-bound}-\eqref{eq:est-err-term3-bound}, we obtain the following bound for \eqref{eq:estimation-err-expanded}:
\begin{align}
    &d_{\omega^*(P_r,P_{G_{\hat{\theta}^*}})}(P_r,{P}_{G_{\hat{\theta}^*}})-\inf_{\theta\in\Theta} d_{\omega^*(P_r,P_{G_{\theta}})}(P_r,P_{G_{\theta}}) \nonumber \\
    & \le 2\sup_{\omega \in \Omega} \Big| \mathbb{E}_{X\sim P_r}[\phi(D_{\omega}(X))] - \mathbb{E}_{X\sim \hat{P}_r}[\phi(D_{\omega}(X))] \Big| \nonumber \\
    & \quad + 2\sup_{\omega \in \Omega,\theta \in \Theta} \Big| \mathbb{E}_{X\sim P_{G_{{\theta}}}}[\psi(D_{\omega}(X))] - \mathbb{E}_{X\sim \hat{P}_{G_{{\theta}}}}[\psi(D_{\omega}(X))] \Big| \nonumber \\
    & = 2\sup_{\omega \in \Omega} \Big| \mathbb{E}_{X\sim P_r}[\phi(D_{\omega}(X))] - \frac{1}{n}\sum_{i=1}^n \phi(D_{\omega}(X_i)) \Big| \nonumber \\
    & \quad + 2\sup_{\omega \in \Omega,\theta \in \Theta} \Big| \mathbb{E}_{X\sim P_{G_{{\theta}}}}[\psi(D_{\omega}(X))] - \frac{1}{m}\sum_{j=1}^m \psi(D_{\omega}(X_j)) \Big|.
    \label{eq:est-err-proof-bound}
\end{align}
Note that \eqref{eq:est-err-proof-bound} is exactly the same bound as that in \cite[Equation~(35)]{kurri2022alphaGAN-extended}. Hence, the remainder of the proof follows from the proof of \cite[Theorem 3]{kurri2022alphaGAN-extended}, where $\alpha=\alpha_G$.

\section{Proof of Theorem \ref{thm:est-error-lower-bound-alpha-infinity}}
\label{appendix:est-error-lower-bound-alpha-infinity}
Let $\phi(\cdot)=-\ell_{\alpha}(1,\cdot)$ and consider the following modified version of $d^{\ell_\alpha}_{\mathcal{F}_{nn}}(\cdot,\cdot)$ (defined in \cite[eq. (13)]{kurri-2022-convergence}):
\begin{align*}
&d^{\ell_\alpha}_{\mathcal{F}_{nn}}(P,Q) = \nonumber\\ & \sup_{\omega \in \Omega} \Big (\mathbb{E}_{X\sim P}[\phi(D_\omega(X))]+\mathbb{E}_{X\sim Q}[\phi(1-D_\omega(X))] \Big) -2\phi(1/2),
\label{eq:dfnn-alpha-loss-modified}
\end{align*}
where
\[D_\omega(x)= \sigma\left(\mathbf{w}_k^\mathsf{T}r_{k-1}(\mathbf{W}_{d-1}r_{k-2}(\dots r_1(\mathbf{W}_1(x)))\right)\coloneqq \sigma\left(f_\omega(x)\right).\]
Taking $\alpha\to\infty$, we obtain
\begin{align} d^{\ell_\infty}_{\mathcal{F}_{nn}}(P,Q) =\sup_{\omega \in \Omega} \Big (\mathbb{E}_{X\sim P}[D_\omega(X)]-\mathbb{E}_{X\sim Q}[D_\omega(X)] \Big).
\label{eq:dfnn-infinity--alpha-loss-modified}
\end{align}
We first prove that $d^{\ell_\infty}_{\mathcal{F}_{nn}}$ is a semi-metric.
\\\noindent \textbf{Claim 1:} For any distribution pair $(P,Q)$, $d^{\ell_\infty}_{\mathcal{F}_{nn}}(P,Q)\ge0$.
\begin{proof}\let\qed\relax
Consider a discriminator which always outputs 1/2, i.e., $D_\omega(x)=1/2$ for all $x$. Note that such a neural network discriminator exists, as setting $\mathbf{w}_k=0$ results in $D_\omega(x)=\sigma(0)=0$. For this discriminator, the objective function in \eqref{eq:dfnn-infinity--alpha-loss-modified} evaluates to $1/2-1/2=0$. Since $d^{\ell_\infty}_{\mathcal{F}_{nn}}$ is a supremum over all discriminators, we have $d^{\ell_\infty}_{\mathcal{F}_{nn}}(P,Q)\ge0$.
\end{proof}
\noindent \textbf{Claim 2:} For any distribution pair $(P,Q)$, $d^{\ell_\infty}_{\mathcal{F}_{nn}}(P,Q)=d^{\ell_\infty}_{\mathcal{F}_{nn}}(Q,P)$.
\begin{proof}\let\qed\relax
\begin{align*}
&d^{\ell_\infty}_{\mathcal{F}_{nn}}(P,Q)  \\
&=\sup_{\omega \in \Omega} \Big (\mathbb{E}_{X\sim P}[D_\omega(X)]-\mathbb{E}_{X\sim Q}[D_\omega(X)] \Big)  \\
& =\sup_{\mathbf{W}_1,\dots,\mathbf{w}_k} \Big (\mathbb{E}_{X\sim P}[D_\omega(X)]-\mathbb{E}_{X\sim Q}[D_\omega(X)] \Big) \\
& \overset{(i)}{=}\sup_{\mathbf{W}_1,\dots,-\mathbf{w}_k} \Big (\mathbb{E}_{X\sim P}[\sigma\left(-f_\omega(x)\right)]-\mathbb{E}_{X\sim Q}[\sigma\left(-f_\omega(x)\right)] \Big) \\
&\overset{(ii)}{=}\sup_{\mathbf{W}_1,\dots,\mathbf{w}_k} \Big (\mathbb{E}_{X\sim P}[1-\sigma\left(f_\omega(x)\right)]-\mathbb{E}_{X\sim Q}[1-\sigma\left(f_\omega(x)\right)] \Big)
 \\
&=\sup_{\mathbf{W}_1,\dots,\mathbf{w}_k} \Big (\mathbb{E}_{X\sim Q}[\sigma\left(f_\omega(x)\right)]-\mathbb{E}_{X\sim P}[\sigma\left(f_\omega(x)\right)] \Big) \\
& = d^{\ell_\infty}_{\mathcal{F}_{nn}}(Q,P),
\end{align*}
where $(i)$ follows from replacing $\mathbf{w}_k$ with $-\mathbf{w}_k$ and $(ii)$ follows from the sigmoid property $\sigma(-x)=1-\sigma(x)$ for all $x$.
\end{proof}
\noindent \textbf{Claim 3:} For any distribution $P$, $d^{\ell_\infty}_{\mathcal{F}_{nn}}(P,P)=0$.
\begin{proof}\let\qed\relax
\begin{align*}
&d^{\ell_\infty}_{\mathcal{F}_{nn}}(P,P)  =\sup_{\omega \in \Omega} \Big (\mathbb{E}_{X\sim P}[D_\omega(X)]-\mathbb{E}_{X\sim P}[D_\omega(X)] \Big)=0.
\end{align*}
\end{proof}
\noindent \textbf{Claim 4:} For any distributions $P,Q,R$, $d^{\ell_\infty}_{\mathcal{F}_{nn}}(P,Q)\le d^{\ell_\infty}_{\mathcal{F}_{nn}}(P,R)+d^{\ell_\infty}_{\mathcal{F}_{nn}}(R,Q)$.
\begin{proof}\let\qed\relax
\begin{align*}
&d^{\ell_\infty}_{\mathcal{F}_{nn}}(P,Q)  \\
&=\sup_{\omega \in \Omega} \Big (\mathbb{E}_{X\sim P}[D_\omega(X)]-\mathbb{E}_{X\sim Q}[D_\omega(X)] \Big)  \\
& =\sup_{\omega \in \Omega} \Big (\mathbb{E}_{X\sim P}[D_\omega(X)] - \mathbb{E}_{X\sim R}[D_\omega(X)]  \\
&\qquad \qquad \quad + \mathbb{E}_{X\sim R}[D_\omega(X)]-\mathbb{E}_{X\sim Q}[D_\omega(X)] \Big)  \\
& \le\sup_{\omega \in \Omega} \Big (\mathbb{E}_{X\sim P}[D_\omega(X)] - \mathbb{E}_{X\sim R}[D_\omega(X)] \Big) \\
&\qquad +  \sup_{\omega \in \Omega} \Big (\mathbb{E}_{X\sim R}[D_\omega(X)]-\mathbb{E}_{X\sim Q}[D_\omega(X)] \Big)  \\
& = d^{\ell_\infty}_{\mathcal{F}_{nn}}(P,R)+d^{\ell_\infty}_{\mathcal{F}_{nn}}(R,Q).
\end{align*}
\end{proof}

Thus, $d^{\ell_\infty}_{\mathcal{F}_{nn}}$ is a semi-metric. The remaining part of the proof of the lower bound follows along the same lines as that of \cite[Theorem 2]{JiZL21} by an application of Fano's inequality \cite[Theorem 2.5]{tsybakov2008nonparametric} (that requires the involved divergence measure to be a semi-metric), replacing $d_{\mathcal{F}_{nn}}$ with $d^{\ell_\infty}_{\mathcal{F}_{nn}}$ and noting that the additional sigmoid activation function after the last layer in the discriminator satisfies the monotonicity assumption so that
$C(\mathcal{P}(\mathcal{X}))>0$ (for $C(\mathcal{P}(\mathcal{X}))$ defined in \eqref{eq:est-error-lower-bound-constant}).

\section{Additional Experimental Results}
\label{appendix:experimental-details-results}

\subsection{Brief Overview of LSGAN}
\label{subsec:lsgan-overview}
The Least Squares GAN (LSGAN) is a dual-objective min-max game introduced in \cite{Mao_2017_LSGAN}. The LSGAN objective functions, as the name suggests, involve squared loss functions for D and G which are written as
\begin{align}
    \sup_{\omega\in\Omega} \; \frac{1}{2}\Big(\mathbb{E}_{X\sim P_r}[(D_\omega(X)-b)^2]+\mathbb{E}_{X\sim P_{G_\theta}}[(D_\omega(X)-a)^2]\Big) \nonumber \\
    \inf_{\theta\in\Theta} \; \frac{1}{2}\Big(\mathbb{E}_{X\sim P_r}[(D_\omega(X)-c)^2]+\mathbb{E}_{X\sim P_{G_\theta}}[(D_\omega(X)-c)^2]\Big).
    \label{eq:lsgan-objectives}
\end{align}
The parameters $a$, $b$, and $c$ are chosen such that \eqref{eq:lsgan-objectives} reduces to minimizing the Pearson $\chi^2$-divergence between $P_r+P_{G_\theta}$ and $2P_{G_\theta}$. As done in the original paper \cite{Mao_2017_LSGAN}, we use $a=1$, $b=0$ and $c=1$ for our experiments to make fair comparisons. The authors refer to this choice of parameters as the 0-1 binary coding scheme. 

\subsection{2D Gaussian Mixture Ring}
\label{appendix:2d-ring}

In Tables \ref{table:2d-ring-sat-success-rates} and \ref{table:2d-ring-sat-failure-rates}, we
report the success (8/8 mode coverage) and failure (0/8 mode coverage) rates over 200 seeds for a grid of $(\alpha_{D}, \alpha_{G})$ combinations for the  \emph{saturating} setting. Compared to the vanilla GAN performance, we find that tuning $\alpha_{D}$ below 1 leads to a greater success rate and lower failure rate. However, in this saturating loss setting, we find that tuning $\alpha_{G}$ away from 1 has no significant impact on GAN performance.

\begin{table}[ht]
\caption{Success rates for 2D-ring with the Saturating $(\alpha_{D}, \alpha_{G})$-GAN over 200 seeds, with top 4 combinations emboldened.
}
\label{table:2d-ring-sat-success-rates}
\renewcommand{\arraystretch}{1.2}
\centering
\begin{tabular}{cl|llllll|}
\cline{3-8}
\multicolumn{2}{c|}{\multirow{2}{*}{\begin{tabular}[c]{@{}c@{}}\% of success \\ (8/8 modes)\end{tabular}}} & \multicolumn{6}{c|}{$\alpha_{D}$} \\ \cline{3-8} 
\multicolumn{2}{c|}{} & 0.5 & 0.6 & 0.7 & 0.8 & 0.9 & 1.0 \\ \hline
\multicolumn{1}{|c|}{\multirow{4}{*}{$\alpha_{G}$}} & 0.9 & \multicolumn{1}{l|}{73} & \multicolumn{1}{l|}{79} & \multicolumn{1}{l|}{69} & \multicolumn{1}{l|}{60} & \multicolumn{1}{l|}{46} & 34 \\ \cline{3-8} 
\multicolumn{1}{|c|}{} & 1.0 & \multicolumn{1}{l|}{\textbf{80}} & \multicolumn{1}{l|}{\textbf{79}} & \multicolumn{1}{l|}{74} & \multicolumn{1}{l|}{68} & \multicolumn{1}{l|}{54} & 47 \\ \cline{3-8} 
\multicolumn{1}{|c|}{} & 1.1 & \multicolumn{1}{l|}{\textbf{79}} & \multicolumn{1}{l|}{77} & \multicolumn{1}{l|}{68} & \multicolumn{1}{l|}{70} & \multicolumn{1}{l|}{59} & 47 \\ \cline{3-8} 
\multicolumn{1}{|c|}{} & 1.2 & \multicolumn{1}{l|}{\textbf{75}} & \multicolumn{1}{l|}{74} & \multicolumn{1}{l|}{71} & \multicolumn{1}{l|}{65} & \multicolumn{1}{l|}{57} & 46 \\ \hline
\end{tabular}
% \end{table}
\vspace*{10pt}
% \begin{table}[ht]
\caption{Failure rates for 2D-ring with the Saturating $(\alpha_{D}, \alpha_{G})$-GAN over 200 seeds, with top 3 combinations emboldened.
}
\label{table:2d-ring-sat-failure-rates}
\renewcommand{\arraystretch}{1.2}
\centering
\begin{tabular}{cl|llllll|}
\cline{3-8}
\multicolumn{2}{c|}{\multirow{2}{*}{\begin{tabular}[c]{@{}c@{}}\% of failure \\ (0/8 modes)\end{tabular}}} & \multicolumn{6}{c|}{$\alpha_{D}$} \\ \cline{3-8} 
\multicolumn{2}{c|}{} & 0.5 & 6 & 7 & 0.8 & 0.9 & 1.0 \\ \hline
\multicolumn{1}{|c|}{\multirow{4}{*}{$\alpha_{G}$}} & 0.9 & \multicolumn{1}{l|}{11} & \multicolumn{1}{l|}{10} & \multicolumn{1}{l|}{12} & \multicolumn{1}{l|}{13} & \multicolumn{1}{l|}{29} & 49 \\ \cline{3-8} 
\multicolumn{1}{|c|}{} & 1.0 & \multicolumn{1}{l|}{\textbf{5}} & \multicolumn{1}{l|}{\textbf{5}} & \multicolumn{1}{l|}{7} & \multicolumn{1}{l|}{8} & \multicolumn{1}{l|}{16} & 30 \\ \cline{3-8} 
\multicolumn{1}{|c|}{} & 1.1 & \multicolumn{1}{l|}{7} & \multicolumn{1}{l|}{9} & \multicolumn{1}{l|}{13} & \multicolumn{1}{l|}{12} & \multicolumn{1}{l|}{13} & 26 \\ \cline{3-8} 
\multicolumn{1}{|c|}{} & 1.2 & \multicolumn{1}{l|}{9} & \multicolumn{1}{l|}{\textbf{5}} & \multicolumn{1}{l|}{9} & \multicolumn{1}{l|}{12} & \multicolumn{1}{l|}{17} & 31 \\ \hline
\end{tabular}
% \end{table}
\vspace*{10pt}
% \begin{table}[h!]
\caption{Success rates for 2D-ring with the NS $(\alpha_{D}, \alpha_{G})$-GAN over 200 seeds, with top 5 combinations emboldened.
%Results for $(\alpha_{D}, \alpha_{G})$-GAN in the NS setting on the 2D-ring dataset, averaged over 200 seeds with top 5 combinations emboldened.
}
\label{table:2d-ring-ns-success-rates}
\renewcommand{\arraystretch}{1.2}
\centering
\begin{tabular}{cl|cccccccc|}
\cline{3-10}
\multicolumn{2}{c|}{\multirow{2}{*}{\begin{tabular}[c]{@{}c@{}}\% of success \\ (8/8 modes)\end{tabular}}} & \multicolumn{8}{c|}{$\alpha_{D}$} \\ \cline{3-10} 
\multicolumn{2}{c|}{} & \multicolumn{1}{l}{0.5} & \multicolumn{1}{l}{0.6} & \multicolumn{1}{l}{0.7} & \multicolumn{1}{l}{0.8} & \multicolumn{1}{l}{0.9} & \multicolumn{1}{l}{1.0} & \multicolumn{1}{l}{1.1} & \multicolumn{1}{l|}{1.2} \\ \hline
\multicolumn{1}{|c|}{\multirow{6}{*}{$\alpha_{G}$}} & 0.8 & \multicolumn{1}{c|}{35} & \multicolumn{1}{c|}{24} & \multicolumn{1}{c|}{19} & \multicolumn{1}{c|}{19} & \multicolumn{1}{c|}{14} & \multicolumn{1}{c|}{16} & \multicolumn{1}{c|}{18} & 10 \\ \cline{3-10} 
\multicolumn{1}{|c|}{} & 0.9 & \multicolumn{1}{c|}{\textbf{39}} & \multicolumn{1}{c|}{37} & \multicolumn{1}{c|}{19} & \multicolumn{1}{c|}{22} & \multicolumn{1}{c|}{16} & \multicolumn{1}{c|}{20} & \multicolumn{1}{c|}{19} & 21 \\ \cline{3-10} 
\multicolumn{1}{|c|}{} & 1.0 & \multicolumn{1}{c|}{34} & \multicolumn{1}{c|}{35} & \multicolumn{1}{c|}{29} & \multicolumn{1}{c|}{28} & \multicolumn{1}{c|}{26} & \multicolumn{1}{c|}{22} & \multicolumn{1}{c|}{20} & 32 \\ \cline{3-10} 
\multicolumn{1}{|c|}{} & 1.1 & \multicolumn{1}{c|}{\textbf{40}} & \multicolumn{1}{c|}{36} & \multicolumn{1}{c|}{31} & \multicolumn{1}{c|}{22} & \multicolumn{1}{c|}{24} & \multicolumn{1}{c|}{15} & \multicolumn{1}{c|}{23} & 25 \\ \cline{3-10} 
\multicolumn{1}{|c|}{} & 1.2 & \multicolumn{1}{c|}{\textbf{45}} & \multicolumn{1}{c|}{38} & \multicolumn{1}{c|}{34} & \multicolumn{1}{c|}{25} & \multicolumn{1}{c|}{26} & \multicolumn{1}{c|}{28} & \multicolumn{1}{c|}{20} & 22 \\ \cline{3-10} 
\multicolumn{1}{|c|}{} & 1.3 & \multicolumn{1}{c|}{\textbf{44}} & \multicolumn{1}{c|}{\textbf{39}} & \multicolumn{1}{c|}{26} & \multicolumn{1}{c|}{28} & \multicolumn{1}{c|}{28} & \multicolumn{1}{c|}{25} & \multicolumn{1}{c|}{31} & 29 \\ \hline
\end{tabular}

\end{table}

In Table \ref{table:2d-ring-ns-success-rates}, we detail the success rates for the NS setting. We note that for this dataset, no failures, and therefore, no vanishing/exploding gradients, occurred in the NS setting. In particular, we find that the $(0.5,1.2)$-GAN doubles the success rate of the vanilla $(1,1)$-GAN, which is more susceptible to mode collapse as illustrated in Figure \ref{fig:NS}. We also find that LSGAN achieves a success rate of 32.5\%, which is greater than vanilla GAN but less than the best-performing $(\alpha_{D}, \alpha_{G})$-GAN.

\subsection{Stacked MNIST}
\label{appendix:stacked-mnist}

For the Stacked MNIST dataset, the discriminator and generator architectures we use are outlined in Tables \ref{table:disc} and \ref{table:gen}, respectively. Both involve four CNN layers whose parameters, include kernel size (i.e., the size of the filter which we denote by Kernel), stride (the number of pixels that the filter moves by), and the output activation functions for each layer. We assume zero padding. Finally, BN in Tables \ref{table:disc} and \ref{table:gen} refers to batch normalization, a technique of normalizing the inputs to each layer where the normalization is over a batch of samples used to train the model at any time. This approach is common in deep learning to avoid cumulative floating point errors and overflows and keep all features in the same range, thereby serving as a computational tool to avoid vanishing and/or exploding gradients. 
% \balance
\begin{table}[t]
\caption{Discriminator architecture for Stacked MNIST.\\ The final sigmoid activation layer is removed for LSGAN.}
\label{table:disc}
\renewcommand{\arraystretch}{1.2}
\centering
\begin{tabular}{|l|l|l|l|l|l|}
\hline
Layer & Output size & Kernel & Stride & BN & Activation \\ \hline
Input & $3 \times 28 \times 28$ &  &  &  &  \\
Convolution & $8 \times 14 \times 14$ & $3 \times 3$ & 2 & Yes & LeakyReLU \\
Convolution & $16 \times 7 \times 7$ & $3 \times 3$ & 2 & Yes & LeakyReLU \\
Convolution & $32 \times 3 \times 3$ & $3 \times 3$ & 2 & Yes & LeakyReLU \\
Convolution & $1 \times 1 \times 1$ & $3 \times 3$ & 2 &  & Sigmoid \\ \hline
\end{tabular}
% \end{table}
% \begin{table}[h]
\vspace*{10pt}
\caption{Generator architecture for the Stacked MNIST experiment.}
\label{table:gen}
\renewcommand{\arraystretch}{1.2}
\centering
\begin{tabular}{|l|l|l|l|l|l|}
\hline
Layer & Output size & Kernel & Stride & BN & Activation \\ \hline
Input & $100 \times 1 \times 1$ &  &  &  &  \\
ConvTranspose & $32 \times 3 \times 3$ & $3 \times 3$ & 2 & Yes & ReLU \\
ConvTranspose & $16 \times 7 \times 7$ & $3 \times 3$ & 2 & Yes & ReLU \\
ConvTranspose & $8 \times 14 \times 14$ & $3 \times 3$ & 2 & Yes & ReLU \\
ConvTranspose & $3 \times 28 \times 28$ & $3 \times 3$ & 2 &  & Tanh \\ \hline
\end{tabular}
% \end{table}
\vspace*{10pt}
% \begin{table}[h]
\centering
\caption{Mean mode coverage reported over 100 seeds for $(\alpha_{D}, \alpha_{G})$-GAN trained on Stacked MNIST with a learning rate of $10^{-3}$ for 50 epochs. The best results are shown in bold.}
\label{table:stacked-grid-1}
\renewcommand{\arraystretch}{1.2}
{%
\begin{tabular}{cc|cccc|}
\cline{3-6}
\multicolumn{2}{c|}{\multirow{2}{*}{\begin{tabular}[c]{@{}c@{}}Mode \\ coverage\end{tabular}}} & \multicolumn{4}{c|}{$\alpha_{D}$} \\ \cline{3-6} 
\multicolumn{2}{c|}{} & 0.9 & 1 & 1.1 & 1.2 \\ \hline
\multicolumn{1}{|c|}{\multirow{4}{*}{$\alpha_{G}$}} & 1 & \multicolumn{1}{c|}{502} & \multicolumn{1}{c|}{541} & \multicolumn{1}{c|}{480} & 508 \\ \cline{3-6} 
\multicolumn{1}{|c|}{} & 1.2 & \multicolumn{1}{c|}{619} & \multicolumn{1}{c|}{586} & \multicolumn{1}{c|}{580} & 598 \\ \cline{3-6} 
\multicolumn{1}{|c|}{} & 1.5 & \multicolumn{1}{c|}{648} & \multicolumn{1}{c|}{684} & \multicolumn{1}{c|}{\textbf{689}} & 645 \\ \cline{3-6} 
\multicolumn{1}{|c|}{} & 2 & \multicolumn{1}{c|}{676} & \multicolumn{1}{c|}{\textbf{690}} & \multicolumn{1}{c|}{\textbf{703}} & \textbf{685} \\ \hline
\end{tabular}%
}
% \end{table}
\vspace*{10pt}
% \begin{table}[h]
\centering
\caption{Mean mode coverage reported over 100 seeds for $(\alpha_{D}, \alpha_{G})$-GAN trained on Stacked MNIST with a learning rate of $5 \times 10^{-4}$ for 100 epochs. The best results are shown in bold.}
\label{table:stacked-grid-2}
\renewcommand{\arraystretch}{1.2}
{%
\begin{tabular}{cc|cc|}
\cline{3-4}
\multicolumn{2}{c|}{\multirow{2}{*}{\begin{tabular}[c]{@{}c@{}}Mode\\coverage\end{tabular}}} & \multicolumn{2}{c|}{$\alpha_{D}$} \\ \cline{3-4} 
\multicolumn{2}{c|}{} & 1 & 2 \\ \hline
\multicolumn{1}{|c|}{\multirow{2}{*}{$\alpha_{G}$}} & 1 & \multicolumn{1}{c|}{665} & 645 \\ \cline{3-4} 
\multicolumn{1}{|c|}{} & 2 & \multicolumn{1}{c|}{693} & \textbf{724} \\ \hline
\end{tabular}%
}
\end{table}

In the main document, we demonstrated the dependence of the computed mode coverage on both the learning rate and the number of training epochs. We now illustrate a commonly used metric for evaluating the quality of the synthetic data, namely, the Frechét Inception Distance (FID). In theory, the FID quantifies the 2-Wasserstein distance between the two distributions, and thus, it is desirable to achieve small values for the FID. In practice, FID %allows comparing the distribution of generated images with that of the real images 
is computed using a lower dimensional latent space for both the real and synthetic images, preferably at a layer close to the output layer. The InceptionNet-V3 deep learning model \cite{heusel2017fid} is used to extract such low-dimensional latent features and use the mean and variance of the features at that layer to compute the FID. 

\newpage In Fig. \ref{fig:stacked-fids}(a) and \ref{fig:stacked-fids}(b), we plot the FID as a function of the learning rate and the number of epochs, respectively. For each such plot, we compare the FID scores for the vanilla GAN ($\alpha_D=\alpha_G=1)$ and LSGAN against different ($\alpha_D=\alpha_G>1)$ values.
%test $(\alpha_{D}, \alpha_{G})$-GAN and LSGAN on Stacked MNIST with varied learning rates (a) and number of epochs (b). 
For these plots, note that we set $\alpha_{D} = \alpha_{G}$. Our motivation for doing so is based on the results shown in Table \ref{table:stacked-grid-1} and \ref{table:stacked-grid-2}, where Table \ref{table:stacked-grid-1} captures the mode coverage for a learning rate of $10^{-3}$ and over 50 training epochs and Table \ref{table:stacked-grid-2} captures the mode coverage for a learning rate of $5\times10^{-4}$ and over 100 training epochs. Our results consistently suggest that $\alpha_{G}$ has a larger impact on the GAN mode coverage performance than $\alpha_{D}$. For both of the abovementioned hyperparameter choices, our results show that $\alpha_{G} = 2$ achieves a wide mode coverage no matter the choice of $\alpha_{D} > 1$; thus, we simplify the $(\alpha_{D}, \alpha_{G})$ search by setting $\alpha_{D} = \alpha_{G}$. Higher values for $\alpha_{D}$ and $\alpha_{G}$ work to mitigate gradient explosion as the derivative of $\alpha$-loss ($\ell_{a}(x)$) approaches 1 as $x \rightarrow 0$ and $\alpha \rightarrow \infty$.

We observe that for smaller values of the learning rate, the FID scores are similar across the GANs; interestingly, we observe a similar trend for lower number of epochs. However, when we increase the learning rate or the number of epochs, the FIDs for vanilla (i.e., $(1,1)$-GAN) and LSGAN increase at a much greater rate than those of the $(\alpha_{D}, \alpha_{G} > 1)$-GANs. These results show that tuning $\alpha_{D}$ and $\alpha_{G}$ above 1 can desensitize the GAN training to hyperparameter initialization, which is particularly desirable when evaluating GANs without prior mode knowledge, as is often the case in practice.

\begin{figure}[t]
    \centering
    \footnotesize
\setlength{\tabcolsep}{1pt}
\begin{tabular}{@{}cc@{}}
  \includegraphics[width=4.5cm]{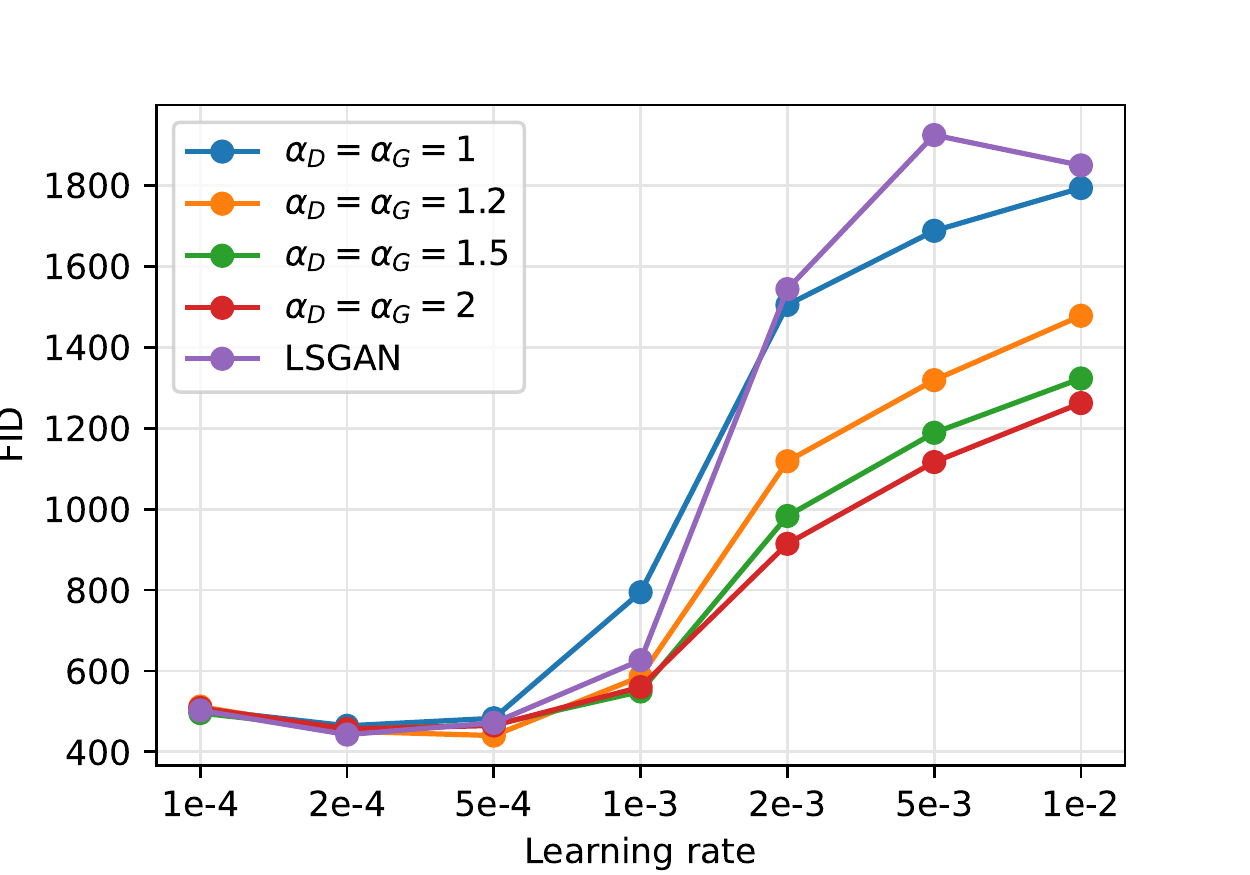}  & {\includegraphics[width=4.5cm]{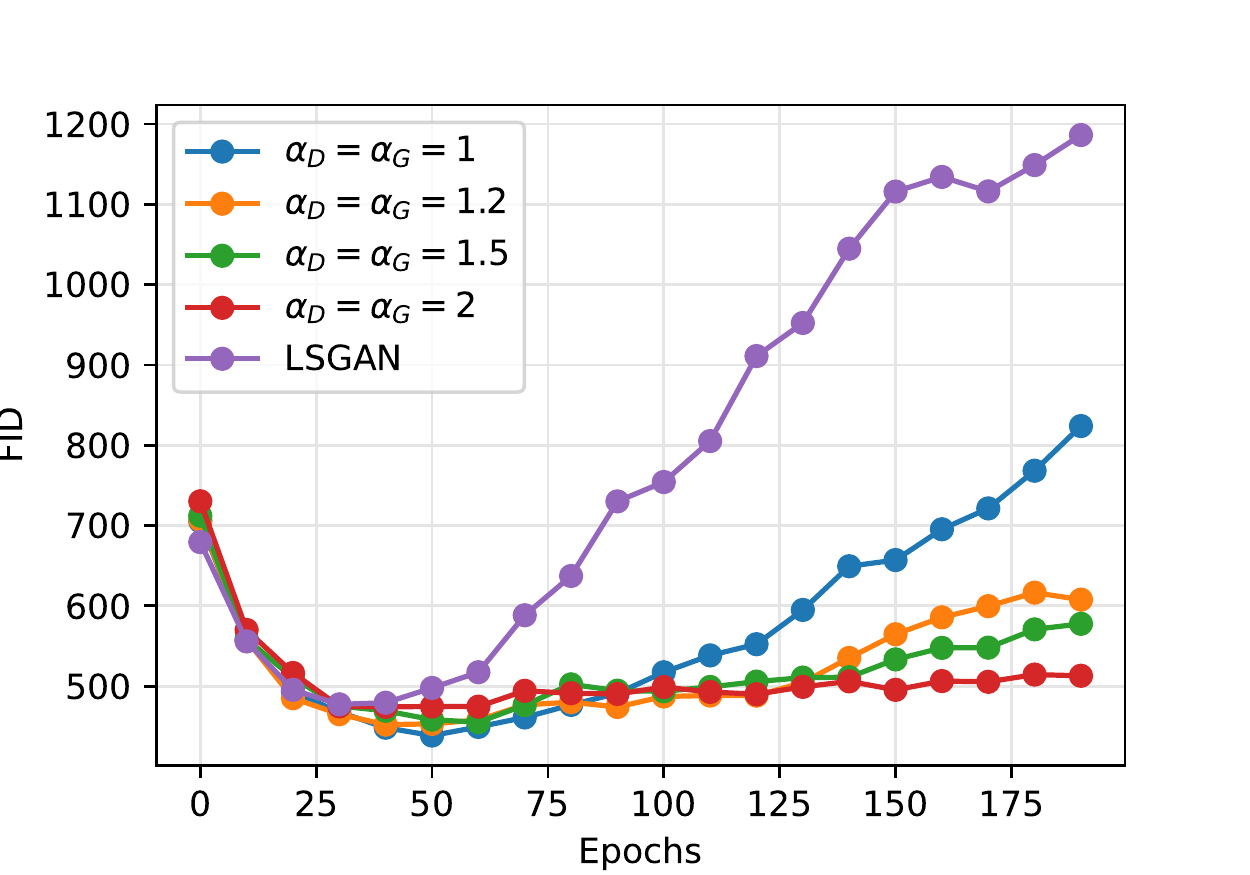}} \\
   (a)  & (b)
\end{tabular}
\caption{FID vs. (a) varied learning rates with fixed epoch numbers ($=50$) and (b) varied epoch numbers with fixed learning rate ($=5\times 10^{-4}$) for different GANs, underscoring the vanilla GAN's hyperparameter sensitivity.
}
\label{fig:stacked-fids}
\end{figure}
\fi
%We remark that the Stacked MNIST dataset is still a relatively easy dataset to learn for all models. We hope to 

% The FID score is designed to extracting `modes in the latent space' with an understanding that for high-dimensional data, a lower dimensional latent feature representation captures the key modes. The Inception-V3 deep learning model used to do so is not trained on our specific dataset, and therefore, despite its wide usage, is not a perfect metric. In fact, comparing Figs. \ref{fig:stacked-output}(b) and \ref{fig:stacked-fids}(b), we observe a discrepancy between the epoch at which the highest mode coverage is achieved (around epoch 10) and that at which the lowest FID is achieved (around epoch 50) for all GAN models. Since the mode coverage statistics is computed using a classifier trained on the MNIST, it provides a window directly to the performance of the GANs considered for this dataset.  

% In practice, it may not be possible to have access to such pretrained classifiers for other datasets. Similarly, limited knowledge also restricts the precise choice of hyperparameters such as learning rate and number of epochs. Using a tunable GAN such as a simple two parameter $(\alpha_D,\alpha_G)$ promises robustness to ranges of these parameters. 

\end{document}